\def\year{2022}\relax

\documentclass[letterpaper]{article} 
\usepackage{aaai22}  
\usepackage{times}  
\usepackage{helvet}  
\usepackage{courier}  
\usepackage[hyphens]{url}  
\usepackage{graphicx} 
\usepackage{natbib}  
\usepackage{caption} 
\DeclareCaptionStyle{ruled}{labelfont=normalfont,labelsep=colon,strut=off} 
\usepackage{amsmath}
\usepackage{amssymb}
\usepackage{amsthm}
\usepackage{graphicx}
\usepackage[noend]{algpseudocode}
\usepackage{algorithm}
\usepackage{bbm}
\usepackage{booktabs}
\usepackage{todonotes}
\usepackage{url}

\usepackage{booktabs}
\usepackage{multirow}
\usepackage{subfig}

\usepackage{pgfplots,pgfplotstable}
\usepgfplotslibrary{groupplots}
\pgfplotsset{compat=newest}
\usepackage{tikz}
\usetikzlibrary{matrix,decorations.pathreplacing, calc, positioning}

\DeclareMathOperator*{\argmin}{arg\,min}

\newtheorem{theorem}{Theorem}

\definecolor{col1}{RGB}{17, 119, 51}
\definecolor{col2}{RGB}{204, 102, 119}
\definecolor{col3}{RGB}{170, 68, 153}
\definecolor{col4}{RGB}{136, 204, 238}
\definecolor{col5}{RGB}{51, 34, 136}
\definecolor{col7}{RGB}{221, 204, 119} 
\definecolor{col6}{RGB}{170, 68, 153} 
\definecolor{col6}{RGB}{153, 153, 51} 
\definecolor{col8}{RGB}{68, 170, 153} 
\definecolor{col9}{RGB}{136, 34, 85}
    
\begin{document}



%
\title{
Shrub Ensembles for Online Classification
}

\author{
    Sebastian Buschjäger\textsuperscript{\rm 1},
    Sibylle Hess\textsuperscript{\rm 2},
    Katharina Morik\textsuperscript{\rm 1}
}
\affiliations{
    \textsuperscript{\rm 1} Artificial Intelligence Group, TU Dortmund, Germany, 
    \{sebastian.buschjaeger, katharina.morik\}@tu-dortmund.de \\
    \textsuperscript{\rm 2} Data Mining Group, Technische Universiteit Eindhoven, Eindhoven, the Netherlands, 
    s.c.hess@tue.nl
%
}



\maketitle

\begin{abstract}
Online learning algorithms have become a ubiquitous tool in the machine learning toolbox and are frequently used in small, resource-constraint environments. Among the most successful online learning methods are Decision Tree (DT) ensembles. DT ensembles provide excellent performance while adapting to changes in the data, but they are not resource efficient. 
Incremental tree learners keep adding new nodes to the tree but never remove old ones increasing the memory consumption over time. Gradient-based tree learning, on the other hand, requires the computation of gradients over the entire tree which is costly for even moderately sized trees.
In this paper, we propose a novel memory-efficient online classification ensemble called shrub ensembles for resource-constraint systems. Our algorithm trains small to medium-sized decision trees on small windows and uses stochastic proximal gradient descent to learn the ensemble weights of these `shrubs'. We provide a theoretical analysis of our algorithm and include an extensive discussion on the behavior of our approach in the online setting. In a series of 2~959 experiments on 12 different datasets, we compare our method against 8  state-of-the-art methods. Our Shrub Ensembles retain an excellent performance even when only little memory is available. We show that SE offers a better accuracy-memory trade-off in 7 of 12 cases, while having a statistically significant better performance than most other methods. Our implementation is available under \url{https://github.com/sbuschjaeger/se-online}.
\end{abstract}

\section{Introduction}
\label{sec:introduction}

Many real-world applications rely on efficient online learning algorithms that are executed on small, resource-constraint devices \cite{daxu/etal/2014}. 
In online learning, the algorithm must process large amounts of data with limited resources in a fast-paced way to provide predictions at any point in time. In addition, data streams often belong to long-running processes which naturally evolve over time and thereby introduce concept drift. Thus, the algorithm must also be able to adapt to new situations and changing data distributions.

Tree ensembles are one of the most popular choices for online learning due to their ability to cope with drift. There are two main strategies to train online tree ensembles. The first strategy uses incremental base-learners such as Hoeffding Trees (HT) \cite{Domingos/2000} or Hoeffding Anytime Trees (HTT) \cite{Manapragada/etal/2018} to incrementally grow individual trees on either different subsets of the data, different sets of features (e.g. as in Online Bagging and its variations \cite{Oza/2005,abdulsalam/etal/2008,gomes/etal/2017b,Gomes/etal/2019}), or on different targets (e.g. as in Online Boosting and its variations \cite{reichler/etala/2004,Oza/2005,kolter/Maloof/2005,pelossof/etal/2009,Chen/etal/2012,Beygelzimer/etal/2015}). The drawback of this strategy is that HT and HTT always keep adding new nodes to the tree without removing old ones. Thus, the size of the trees grows over time which is not suitable for applications on small devices.
The second strategy is to use a fixed tree structure and to update the individual split nodes of the tree via (stochastic) gradient-descent \cite{Irsoy/etal/2012,Kontschieder/etal/2015,Seyedhosseini/etal/2015,Shen/etal/2018}. While this approach uses a fixed amount of memory, it requires the costly computation of gradients by backpropagation through the entire tree, which is not suitable for the processing capabilities of small devices.

In this paper, we establish a third strategy. We propose to maintain a bounded but dynamic ensemble of so-called decision shrubs. 
Just as in botany, decision shrubs are small- to medium-sized trees which compete against each other. Our algorithm trains shrubs on small windows and uses stochastic proximal gradient descent to learn the weights of individual shrubs in the ensemble. Shrubs with sub-optimal performance are aggressively pruned from the ensemble while new shrubs are regularly introduced. This makes our algorithm fast and memory-efficient while it retains a high degree of adaptability to evolving data streams. 
In contrast to incremental learners, our trees never exceed a fixed size. In contrast to gradient-based approaches, we replace costly gradient-computations with a continuous re-training of trees on small batches. Our contributions are as follows:

\begin{itemize}
    \item \textbf{Shrub ensembles:}  We present a learning objective which enforces a small and effective ensemble. The Shrub Ensemble (SE) algorithm uses a variant of the theoretically well-founded proximal gradient descent, which introduces the training of new shrubs on small batches of the data and the automatic removal of sub-optimal trees from the ensemble.
    \item \textbf{Theoretical analysis:} We provide a theoretical analysis of the behavior of our algorithm and discuss under which conditions new shrubs are added and old shrubs are removed from the ensemble. 
    \item \textbf{Experimental analysis:} In a series of 2~959 experiments on 12 different datasets we compare our method against 8 other state-of-the-art methods. Our Shrub Ensembles retain an excellent performance even when only little memory is available. We show that SE offers a better accuracy-memory trade-off in 7 of 12 cases, while having a statistically significant better performance than most other methods.
\end{itemize}

\section{Related Work}
\label{sec:related-work}

Closely related to our approach are online learning algorithms of tree-based ensembles. Here we identify three general research directions: racing-based, gradient-based algorithms and window-based algorithms.

\subsection{Racing-based algorithms} Racing-based algorithms propose multiple hypotheses and let them race each other for the best performance \cite{maron/moore/1993,maron/Moore/1997,loh/etal/2013}. \cite{Domingos/2000} applied this idea to decision tree (DT) induction and proposed the Hoeffding Tree (HT) algorithm. HT is an online version of the regular top-down induction of DTs. The induction starts with the root node and creates a list of split hypotheses. The performance of each hypothesis is measured while new items arrive and are compared against each other using Hoeffding's inequality. Once Hoeffding's bound yields that a split is significantly better than any other split, the split is integrated into the tree. Consecutively, new child nodes are introduced which receive a new list of candidate splits and the race continues. The vanilla HT algorithm is only able to handle categorical input variables and uses the classification error as a split criterion; multiple extensions have been proposed to generalize this approach~\cite{hulten/etal/2001,Holmes/etal/2005,Pfahringer/etal/2007,Pfahringer/etal/2008,bifet/Gavalda/2009,Rutkowski/etal/2013,Mirkhan/etal/2019}. 

Most notably, the overall approach of HTs has been improved by Hoeffding Anytime Trees (HTT) \cite{Manapragada/etal/2018}. HTTs greedily select the best split nodes after a few examples before Hoeffding's bound deems it significantly better than the other splits, but keeps evaluating \emph{all} split candidates in all nodes. Then, it re-orders the entire tree if the initial greedy choice becomes sub-optimal due to Hoeffding's Bound.
Given HT and its siblings, a multitude of different ensemble techniques have been proposed including online bagging variants \cite{kolter/Maloof/2005,Oza/2005,abdulsalam/etal/2008,gomes/etal/2017b,Gomes/etal/2019} and online boosting variations \cite{reichler/etala/2004,Oza/2005,pelossof/etal/2009,Chen/etal/2012,Beygelzimer/etal/2015}. More specialized approaches, which are, e.g., specifically designed for concept drift data are also available~\cite{abdulsalam/etal/2008,Bifet/etal/2010,hoens/etal/2011,gomes/etal/2017b,Gomes/etal/2019}. For an overview see \cite{Gomes/etal/2017a,krawczyk/etal/2017} and references therein. 

\subsection{Gradient-based algorithms} Gradient-based algorithms view the entire DT ensemble as a single model which is trained via (stochastic) gradient descent \cite{Kontschieder/etal/2015,Seyedhosseini/etal/2015,ahmetouglu/etal/2018,Shen/etal/2018,irsoy/ethem/2021}. A single DT is represented by the function 
\begin{equation}  
	h(x) = \sum_{i \in L}g_i(x)\prod_{j \in P_i} s_j(x),
	\label{eq:DTPred}
\end{equation}
where $L$ is the set of leaves, $P_i$ is the path from the root node to the $i$-th leaf node, $s_j \colon \mathbb R^d \to [0,1]$ is a split function and $g_i \colon \mathbb R^d \to \mathbb R^C$ is the leaf's prediction function. The entire ensemble is then given by
$f(x) = \frac{1}{M}\sum_{i=1}^M h_i(x)$
and the objective is to find the function $f$ that minimizes the loss $\ell(f(x), y)$. Regular DTs use axis-aligned splits $s(x) = \mathbbm 1\{x_i \le t\}$ in which $i$ is a pre-computed feature index and $t$ is a threshold. This function is not smooth which makes the optimization via gradient descent difficult. Thus, a \emph{soft} DT with $s(x) = \sigma(z(x))$ (for the children at the right side) and $s(x) = 1 - \sigma(z(x))$ (for the children at the left side) is often used where $\sigma$ is the sigmoid function and $z \colon \mathbb R^d \to \mathbb R$ is another split function. A common example for the split function is a linear function $z(x) = \langle x, w \rangle$. The weight vector $w$ is thereby a part of the optimization objective and determines the features which are relevant in the corresponding split by its nonzero entries. 

Some approaches introduce sparsity regularization terms for $w$ in order to enforce using fewer features in $z$ \cite{Yildiz/etal/2014}. Other approaches apply dropout to the tree edges during training \cite{irsoy/ethem/2021}, add the possibility to resize the tree during learning \cite{Tanno/etal/2019} or map examples into a lower dimensional embedding space for smaller trees~\cite{Kumar/etal/2017}. Lastly, we want to note that soft decision trees can also be viewed as a specialized Deep Learning architecture, which has been explored to some extent \cite{Frosst/Hinton/2017,Biau/etal/2019}. While all these approaches have the advantage that the tree size is (relatively) fixed, they require the costly computation of gradients through backpropagation over the entire tree and suffer from the vanishing gradient problem, which additionally slows down convergence \cite{hochreiter/1998}. 

\subsection{Window-based algorithms}
Highly related to our approach are DTs which are learned on a window of the data.
Training a single DT on a sliding window has already been proposed in the 80s \cite{kubat/1989}. The FLORA method constructs logical formulas of the form $A_1 \land A_2 \land \dots \land A_n \implies B$ using so-called rough-sets \cite{pawlak/1982}. Clearly, the above formula also represents a decision tree, even though the training of such a tree does not follow the more established CART or ID3 algorithm. Street and Kim extend this approach by introducing a heuristic which re-trains individual trees of the ensemble on small batches of the data whenever the performance of a classifier deteriorates\cite{street/kim/2001}. Unfortunately, we could not find any evidence that this simple baseline has been considered much beyond its original publication or the related variants FLORA2 - FLORA4 \cite{widmer/Kubat/1996}. 
More `recent' papers often train (batched) Naive Bayes, SVM or KNN over a fixed-sized window  \cite{scholz/klinkenberg/2005,bifet/gavalda/2007}, while current work does not compare against fixed-sized windows at-all \cite{gomes/etal/2017b,Gomes/etal/2019}. 

Our approach extends the training of DTs over a fixed-sized window to learning an optimal combination of trees on the fly via proximal gradient descent.



\section{Shrub Ensembles}
\label{sec:ensemble-learning}

We start by formalizing our problem setup. Let $\mathcal{S} = \{(x_i,y_i)|i\in\{0,\dots\}\}$ be an open-ended sequence of $d$-dimensional feature vectors $x_i \in \mathbb R^d$ and labels $y_i\in \mathcal Y \subseteq \mathbb R^C$. For classification problems with $C \ge 2$ classes we encode each label as a one-hot vector $y = (0,\dots,0,1,0,\dots,0)$, having $y_c =1$ for the assigned label $c \in \{0,\dots,C-1\}$; for regression problems we have $C=1$ and $\mathcal Y = \mathbb R$. In this paper, we focus on classification problems, even though the presented approach is directly applicable to regression, as well. Let $S[t:t+T] = \{(x_t,y_t),\dots,(x_{t+T},y_{t+T})\}$ denote the sub-sequence of length $T$ starting at element $t$. Our goal is to maintain a suitable model $f \colon \mathbb R^d \to \mathcal Y$, which integrates the knowledge of previously observed examples $S[0:t-1]$, while also offering a good prediction $f(x_t)$ for the next data point $x_t$ \emph{before} the true label $y_t$ is known. There are three crucial challenges in this setting:
\begin{itemize}
    \item \textbf{Computational efficiency:} The algorithm must process examples at least as fast as new examples arrive.
    \item \textbf{Memory efficiency:} The algorithm has only a limited budget of memory and fails if more memory is required.
    \item \textbf{Evolving data streams:} The underlying distribution of the data might change over time, e.g., in the form of concept drift, and the algorithm must adapt to new data trends to preserve its performance.
\end{itemize}

In this paper, we assume that $f$ is an additive ensemble of $K = |\mathcal H|$ base learners from some model class $\mathcal H = \{h \colon \mathbb R^d \to \mathbb R^C\}$, where $K$ is potentially very large or even infinite. Each of the $K$ learners $h_i\in\mathcal{H}$ is associated with a weight $w_i \geq 0$ and the ensemble is given by:
\begin{equation}
f(x) = \sum_{i=1}^K w_i h_i(x)
\end{equation}

Now, assume that the set of models $\mathcal H$ is fixed to a finite set \emph{beforehand} (we will discuss how to adapt  $\mathcal H$ during optimization later). Then our goal is to learn the optimal weights $w_i$ for each base learner. 
Since we require a memory-efficient and adaptable algorithm, only $M \ll K$ trees should receive a nonzero weight and the remaining $K-M$ hypotheses should have a zero weight $w_i = 0$. In this way, the computation of $f(x_t)$ becomes very efficient since only $M$ instead of $K$ models must be executed for prediction. Additionally, we are free to select another set of $M$ hypotheses if there is a drift in the data, thus retaining the adaptability of the algorithm. Formally, we propose the following optimization objective
\begin{equation}
    \begin{split}
  \argmin_{ w \in \mathbb R^K}\ & \sum_{t=1}^T  \ell \left( f_{S[0:t-1]}(x_t), y_t \right)
  \\ 
\text{s.t.}\ & \lVert  w\rVert_0 \le M, w_i \ge 0, \sum_{i=1}^K w_i = 1
\label{eq:objective}
\end{split}
\end{equation}
where $M \ge 1$ is the maximum number of ensemble members, $\ell \colon \mathbb R^C \times \mathcal Y \to \mathbb R_{+}$ is a loss function, $\lVert w\rVert_0 = \sum_{i=1}^K \mathbbm{1}\{w_i \not= 0\}$ is the $0-$norm which counts the number of nonzero entries in $w$ and $f_{S[0:t-1]}\colon \mathbb R^d \to \mathbb R^C$ is the model at time $t$. For concreteness, we now focus on the (multi-class) MSE loss but note that our implementation also supports other loss functions such as the cross-entropy loss:
\begin{equation}
\ell(f_{S[0:t-1]}(x_t), y_t) =  \frac{1}{C}\lVert f_{S[0:t-1]}(x_t) - y_t\rVert^2
\end{equation}


\subsection{Optimizing the weights when $\mathcal H$ is known}
The direct minimization of $L(w,h) = \sum_{t=1}^T \ell(f_{S[0:t-1]}(x_t), y_t) $ is infeasible, since $\mathcal S$ is open-ended and unknown beforehand. However, we can store small batches $\mathcal B = S[t-B:t]$ of the incoming data (e.g. a sliding window) and use them to approximate our objective via the sample mean
\begin{equation}
L_{\mathcal B}(w,h) = \frac{1}{B C}\sum_{(x,y)\in\mathcal{B}}\lVert f(x) - y\rVert^2.
\end{equation}
The function $L$ is convex and smooth, and its global optimum can be easily derived via its stationary points. However, the feasible set
\begin{equation}
\Delta = \left\{w \in \mathbb R_+^K \middle| \sum_{i=1}^K w_i = 1,  \lVert w\rVert_0 = M\right\}
\end{equation}
is not convex, which makes the convex optimization problem to minimize $L(w,h)$ over $w\in\mathbb R^d$ a nonconvex problem, minimizing $L(w,h)$ over the feasible set $w\in\Delta$.

A popular choice to enable the integration of constraints into gradient-based optimization methods is to use proximal gradient descent. In particular, since we consider batch-wise updates of parameters, 
we discuss the application of Stochastic Proximal Gradient Descent (SPGD). 

SPGD is an iterative algorithm, where every iteration consists of two steps: first, a gradient descent update of the objective function $L_{\mathcal B}(w,h)$ is performed without considering any constraint. Then, the prox-operator is applied to project its argument onto the feasible set $\Delta$. 
The proximal gradient update for every iteration is then given as
\begin{equation}
w \gets \mathcal P \left( w - \alpha \nabla_w L_{\mathcal B}(w) \right),
\end{equation}
where $\alpha \in \mathbb R_+$ is the step-size and $\mathcal P \colon \mathbb R^K \to \Delta$ is the corresponding prox-operator. 
For $\nabla_{w} L(w, h) = \nabla_w L_{\mathcal B}(w)$ (e.g. if $\mathcal B = \mathcal S$), we obtain the `regular' proximal gradient descent algorithm. \cite{Kyrillidis/etal/2013} study the computation of the prox-operator for the combination of sparsity requirements and simplex constraints, which define our feasible set. Assuming that the vector $w$ is decreasingly ordered, such that $w_1\geq\ldots \geq w_K $, they present an operator that sets the $K-M$ smallest entries in $w$ to zero and projects the $M$ largest values onto the probability simplex:
\begin{equation}
\begin{split}
\mathcal P\left(w\right)_i &= \begin{cases} 0 &\text{if~} i>M \\ 
[w_i - \tau]_+ &\text{otherwise~} \end{cases} \\ 
\text{where } \tau &= \frac{1}{\beta}\left(\sum_{i=1}^\beta w_i - 1\right), \\ 
\beta &= \max \left\{j \middle| w_j > \frac{1}{j}\sum_{i=1}^j \left(w_i - 1\right), j\le M\right\} 
\end{split}
\end{equation}
\subsection{Optimizing $\mathcal{H}$ simultaneously with the weights}
If the set of classifiers $\mathcal H$ is large, then the computation of the weight-gradient can be costly, even when the prox-operator ensures that only $M$ models receive a nonzero weight. In turn, a small candidate set $\mathcal H$ restricts the possibilities to adapt to the environment such that the model can not adequately react to concept drift. 
A natural solution to this problem is to drop the assumption that all models in $\mathcal H$ are known beforehand and to instead dynamically change $\mathcal H$ with new incoming data. To do so we maintain at most $M$ trees in the ensemble whose corresponding entries in $w$ are nonzero. 

In every iteration we add a new tree to the ensemble and then update the $M+1$ weights of $w$. The update ensures that at least one of the $M+1$ trees obtain a weight of zero, which will be replaced in the subsequent step with a newly trained tree. The new tree will be a part of the ensemble as long as its weight is in subsequent updates not set to zero. 
The trees are trained on small batches and are thus comparably small. 

Our method \emph{Shrub Ensembles} (SE) is outlined in Algorithm~\ref{fig:SE}.
We start with an empty buffer $\mathcal B$ and an empty set of trees $\mathcal H$. For every new data item we update the sliding window buffer (line 3-4), train a new classifier (e.g., via CART) and initialize its weight with zero (line 6-8). Then, we perform the gradient step followed by the prox-operator (line 9-11). Finally, we remove classifiers with a weight of $0$ (line 12). The intuition of our approach is that a newly trained shrub which (significantly) improves the ensemble's prediction will likely receive a large enough weight after the gradient update to survive the subsequent prox-operator. If, however, the tree does not improve the ensemble's prediction much it might only receive  little gradient-mass, such that the tree is removed from the ensemble immediately. 

\begin{algorithm}
	\begin{algorithmic}[1]
	\State{$w \gets (0)$; $\mathcal B \gets [~]; \mathcal H \gets [~]$} \Comment{\parbox[t]{0.42\linewidth}{Init.}}
	\For{next item $(x,y)$}
	   \If{$|\mathcal B| = B$} \Comment{\parbox[t]{0.42\linewidth}{Update batch}}
	        \State{$\mathcal B$.\texttt{pop\_first()}}
	   \EndIf
	   \State{$\mathcal B$.\texttt{append($(x,y)$)}}
        \State{$h_{new} \gets \texttt{train}(\mathcal B)$} \Comment{\parbox[t]{0.42\linewidth}{Add new classifier}}
        \State{$\mathcal H$.\texttt{append($h_{new}$)}}
        \State{$w \gets (w_1,\dots,w_M, 0)$} \Comment{\parbox[t]{0.42\linewidth}{Initialize weight}}
        \State{$w \gets w - \alpha \nabla_w L_{\mathcal B}(w)$} \Comment{\parbox[t]{0.42\linewidth}{Gradient step}}
        \State{$w, \mathcal H \gets \texttt{sorted}(w, \mathcal H)$} \Comment{\parbox[t]{0.42\linewidth}{Sort decreasing order}}
        \State{$w \gets \mathcal P ( w )$} \Comment{\parbox[t]{0.42\linewidth}{Project on feasible set}}
        \State{$w, \mathcal H \gets \texttt{prune}(w,\mathcal H)$} \Comment{\parbox[t]{0.42\linewidth}{Remove zero weights}}
    \EndFor
	\end{algorithmic}
	\caption{Shrub Ensembles.}
	\label{fig:SE}
\end{algorithm}
\subsection{Theoretical performance of Shrub Ensembles}
Theorem \ref{th:learning-shrubs} formalizes the theoretical behavior of Shrub Ensembles. It shows that whenever a new, previously unknown relationship (or concept) between observations and labels is discovered, then SE will include the newly trained tree in the ensemble, given an appropriate choice for the step size. In particular this means two things: first, when $M=1$ then SE resembles the continuous re-training of trees over a sliding window of fixed size, similar to the previously discussed FLORA algorithm. Second, SE will always incorporate a new concept into the ensemble while keeping track of past concepts, only replacing that tree with the smallest contribution to the entire ensemble. For large step sizes our approach is very aggressive as we introduce a new tree immediately in the ensemble when a single new concept arrives. For very fast changing data, this can be beneficial, but in some settings this can hurt the performance, e.g., if the data is very noisy. 

Theorem \ref{th:learning-shrubs} has two crucial assumptions. First, shrubs are assumed to be fully-grown so that they perfectly isolate the points in the current window. Second, the step size must be large enough. Turning this statement around, SE might decide \emph{not} to include a new tree into the ensemble if the step size is smaller than $\frac{BC}{4m}$ or if trees are not perfectly fitting the current batch. In this case, the new concept $(x,y)$ might be ignored if it only appears a few times in the current window. If, however, the new concept appears multiple times in the window a newly trained tree will likely `overfit' this new concept and therefore receive a big enough weight to replace one of the other trees. It follows that for large step sizes $\alpha > \frac{BC}{4m}$ and fully-grown trees SE will follow changes in the distribution very quickly, whereas for smaller step sizes and `weaker' trees it will be more resilient to noise in the data.

\begin{theorem}
\label{th:learning-shrubs}
Let $M \ge 1$ be the maximum ensemble size in the shrub ensembles (SE) algorithm and let $B$ be the buffer size.
Consider a classification problem with $C$ classes.
Further, let $m \le M$ be the number of models in the ensemble. Now assume that a new observation $(x_B,y_B)$ arrives, which was previously unknown to the ensemble so that $\forall j=1,\dots,m \colon h_j(x_B) \not= y_{B}$. Let SE train fully-grown trees with $h_j(x)\in \{0,1\}^C$ and let $h$ be the new tree, trained on the current window, such that $\forall i=1,\dots,B \colon h(x_i) = y_{i}$. Then we have for $\alpha > \frac{BC}{4m}$ the following cases:

\begin{itemize}
    \item (1) If $m < M$, then $h$ is added to the ensemble
    \item (2) If $m = M$, then $h$ replaces the tree with the smallest weight from the ensemble.
\end{itemize}
\end{theorem}
The proof is stated in the appendix.

\subsection{Runtime and Memory of Shrub Ensembles}
The continuous training of new models in SE might seem costly, but for reasonable choices of base learners and window sizes, the training is comparably quick. 
First, the regular CART or ID3 algorithm requires $\mathcal O(d N^2 \log N)$ runtime where $N$ is the number of datapoints. In our case we have $N = B$, that is, the runtime of tree induction is limited by the window size. Second, there are many efficient heuristics and implementations available for decision tree induction that improve the theoretical and practical runtime of tree learning, e.g. by only considering a well-chosen subset of splits \cite{chen/etal/2016,ke/etal/2017,Prokhorenkova/etal/2018,geurts/etal/2006}. 
The computation of the prox-operator in Algorithm \ref{fig:SE} is in $\mathcal O(M \log M)$ \cite{Wang/etal/2013}. The complexity is dominated by the sorting of $w$. To further decrease the runtime we can maintain a sorted list of $w$ and $\mathcal H$ instead of sorting them from scratch (line 10). This can efficiently be done via a binary search tree which only requires $\mathcal O(\log M)$ runtime for the insertion and deletion (line 12) of items. Hence the total complexity of SE is $\mathcal O\left(d B^2 \log B + \log M\right)$.

Regarding the memory consumption, we note that the size of the trained trees are inherently limited by the window size -- regardless of the specific training algorithm. A fully-grown tree that perfectly separates the observations in $\mathcal B$ requires at most $B$ leaf nodes. Therefore, the \emph{total} number of nodes used by a tree is upper-bounded by the window size with $2^{\log_2 B + 1}  - 1 = 2 \cdot B -1$. It follows that Algorithm \ref{fig:SE} stores at most $B$ examples and $M+1$ models, each having at most $2 \cdot B -1$ nodes. This makes our shrub ensembles an overall fast and memory efficient algorithm.

\section{Experiments} 
\label{sec:experiments}

In our experimental evaluation we are interested in the performance of our shrub ensembles in comparison to recent state-of-the-art methods. We are specifically interested in the accuracy-memory trade-off of these methods. For our analysis we adopt a hardware-agnostic view which assumes that we are given a fixed memory budget for our model, which should, naturally, maintain a state-of-the-art performance.
For racing-based algorithms (cf.\@ Section~\ref{sec:related-work}) we use Online Naive Bayes (NB), Hoeffding Trees (HT), Hoeffding Anytime Trees (HTT), Streaming Random Patches (SRP), Adaptive Random Forest (ARF), Online Bagging (Bag) and Smooth Boost (SB) implemented in MOA \cite{Bifet/etal/2010b}. 
For gradient-based approaches (cf.\@ Section~\ref{sec:related-work}) we implemented soft decision tree ensembles (SDT) using PyTorch \cite{Paszke/etal/2018}. For Shrub Ensembles (SE) we used our own \texttt{C++} implementation. We compare the performance of each algorithm using the average test-then-train accuracy and the average model size (in kilobyte) on 12 different datasets depicted in the appendix. Our code is available under \url{https://github.com/sbuschjaeger/se-online}. 

We measure the model size as the entire model, including any stored variables (e.g., including the sliding window). A careful reader might view this comparison as slightly biased against MOA since it is implemented in Java, whereas the other algorithms are implemented in \texttt{C++} (with a Python interface). Unfortunately, there is currently no alternative, efficient MOA implementation available. Preliminary projects to implement MOA in \texttt{C++}\footnote{\url{https://github.com/huawei-noah/streamDM-Cpp}} or Python\footnote{The authors of \url{https://riverml.xyz/} confirmed that they currently strive for functional and `feature complete' code and perform optimizations later on during development.} have not been finalized, yet.
Thus, we put effort in making this comparison fair by computing a reference size of each MOA model first (before it received any data points), which is then subtracted from the measurements. This way, we only account for changes in the model due to new items and do not include the `static' overhead of Java. 

To ensure a fair comparison between the hyperparameter choices of each individual algorithm, we follow the methodology presented in \cite{bergstra/bengio/2012}. In a series of preliminary experiments, we identify reasonable ranges for each hyperparameter and method (e.g., number of trees in an ensemble, window size, step sizes etc.). Then, for each method and dataset we sample at most $50$ random hyperparameter configurations from these ranges and evaluate their performance. 
An example of such a configuration can be found in the appendix and further details can be taken from the source code. To ensure timely and realistic results, we remove each configuration which took longer than two hours to complete or models which exceed $100$ MB in sizes. In summary, we test $312$ different configurations per dataset totaling to $3~744$ experiments of which we analyzed $2~786$ experiments. For the experiments we used a cluster node with 256 AMD EPYC 7742 CPUs and 1TB ram in total. An anonymized version of our source code is available in the appendix, and our source code as well as the additional results will be made public after acceptance.

\subsection{Quantitative Analysis}
As mentioned before, we are interested in the most accurate models with the smallest memory consumption. Clearly these two metrics can contradict each other. Hence
we compute the Pareto front of each method which contains those parameter configurations that are not dominated across one or more dimensions. Then, we summarize the algorithm's performance via the area-under the Pareto front (APF) normalized by the biggest model for the given dataset for comparison. For the individual accuracies of each method please consult the appendix.

\begin{table*}
\centering
\caption{\label{tab:auc_experiment} Normalized area under the Pareto front (APF) for each method and each dataset with models smaller than $100$ MB. Rounded to the fourth decimal digit. Larger is better. The best method is depicted in bold.}

\begin{tabular}{lrrrrrrrrr}
\toprule
 &     ARF &     Bag &      HT &     HTT &      NB &      SB &     SDT &      SE &     SRP \\
\midrule
agrawal\_a  &  0.8259 &  0.8939 &  0.9145 &  0.9136 &  0.7429 &  0.9124 &  0.0977 &  \textbf{0.9355} &  0.8877 \\
agrawal\_g  &  0.7738 &  0.8535 &  0.8601 &  0.8732 &  0.7427 &  0.8759 &  0.4548 &  \textbf{0.9157} &  0.8571 \\
airlines   &  0.5679 &  0.3461 &  0.5654 &  0.6588 &  0.6693 &  0.5048 &  0.3877 &  \textbf{0.6818} &  0.5369 \\
covtype    &  0.9274 &  0.8157 &  0.8487 &  0.8712 &  0.6458 &  0.8196 &  0.0374 &  0.9284 &  \textbf{0.9285} \\
elec       &  0.9081 &  0.8946 &  0.8641 &  0.8771 &  0.7616 &  0.8821 &  0.4251 &  \textbf{0.9401} &  0.9091 \\
gas-sensor &  0.9238 &  0.9120 &  0.8474 &  0.9217 &  0.7287 &  0.7580 &  0.1716 &  \textbf{0.9679} &  0.9537 \\
led\_a      &  0.7149 &  0.7050 &  0.6846 &  0.7095 &  0.6441 &  0.7199 &  0.1011 &  0.7233 &  \textbf{0.7240} \\
led\_g      &  0.7085 &  0.7034 &  0.6825 &  0.7089 &  0.6439 &  0.7174 &  - &  \textbf{0.7144} &  0.7124 \\
nomao      &  0.9793 &  0.9613 &  0.9443 &  0.9668 &  0.9277 &  0.9606 &  0.8863 &  \textbf{0.9866} &  0.9752 \\
rbf\_f      &  0.7550 &  0.5171 &  0.3807 &  0.4014 &  0.2975 &  0.4217 &  0.0725 &  0.7302 &  \textbf{0.7739} \\
rbf\_m      &  \textbf{0.8541} &  0.7566 &  0.6262 &  0.6803 &  0.3295 &  0.6478 &  0.1667 &  0.7962 &  0.8479 \\
weather    &  0.7788 &  0.7827 &  0.7510 &  0.7658 &  0.6958 &  0.7688 &  0.3287 &  0.7586 &  \textbf{0.7895} \\
\bottomrule
\end{tabular}
\end{table*}

\begin{figure}[t!]
\centering
\includegraphics[width=\columnwidth, keepaspectratio]{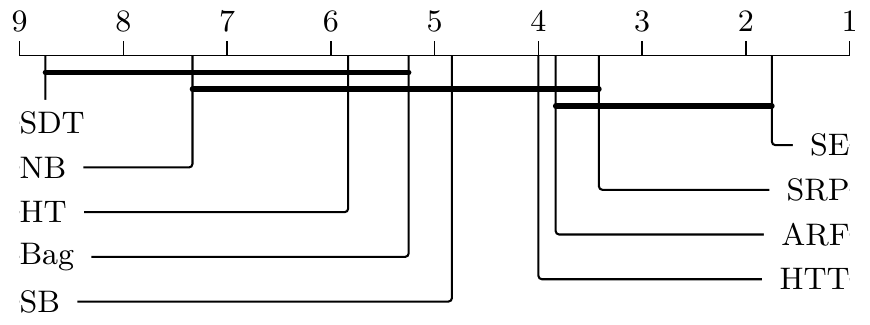}
\caption{Critical Difference Diagram for the normalized area under the Pareto Front for different methods over multiple datasets. For all statistical tests $p=0.95$ was used. More to the right (lower rank) is better. Methods in connected cliques are statistically similar.}
\label{fig:CD}
\end{figure}

Table \ref{tab:auc_experiment} shows the normalized area under the Pareto front. It can be seen that SE and SRP generally performs best followed by ARF. Our SE method ranks first on 7 datasets offering the best accuracy-memory trade-off followed by SRP which ranks first on 4 datasets. On the third place we find ARF which ranks first on the rbf\_m dataset. SDT does not perform well. We hypothesize that this is due to the random initialization combined with the vanishing gradient problem. On the led\_g dataset it did not finish the computation in under 2 hours and thus was removed for our evaluation. To give a statistical meaningful comparison we present the results in Table \ref{tab:auc_experiment} as a CD diagram \cite{demvsar/2006}. In a CD diagram each method is ranked according to its performance and a Friedman-Test is used to determine if there is a statistical difference between the average rank of each method. If this is the case, then a pairwise Wilcoxon-Test between all methods checks whether there is a statistical difference between two classifiers. CD diagrams visualize this evaluation by plotting the average rank of each method on the x-axis and connect all classifiers whose performances are statistically similar via a horizontal bar. Figure \ref{fig:CD} shows the corresponding CD diagram, where $p=0.95$ was used for all statistical tests. It can be seen that SE ranks first with an average rank between $1-2$ with some distance to SRP which -- on average -- ranks between $3-4$th place closely followed by ARF. Next, there is $\{SRP, ARF, HTT, SB, Bag, HT, NB\}$ which form a second clique and $\{Bag, HT, NB, SDT\}$ which forms the last clique. While all three methods $\{SE,SRP,ARF\}$ are in the same clique and hence offer similar performance, SE has some distance. It is only present in this clique meaning that it is statistically better than $\{HTT, SB, Bag, HT, NB, SDT\}$.

\subsection{Qualitative Analysis}
\begin{figure}[t!]
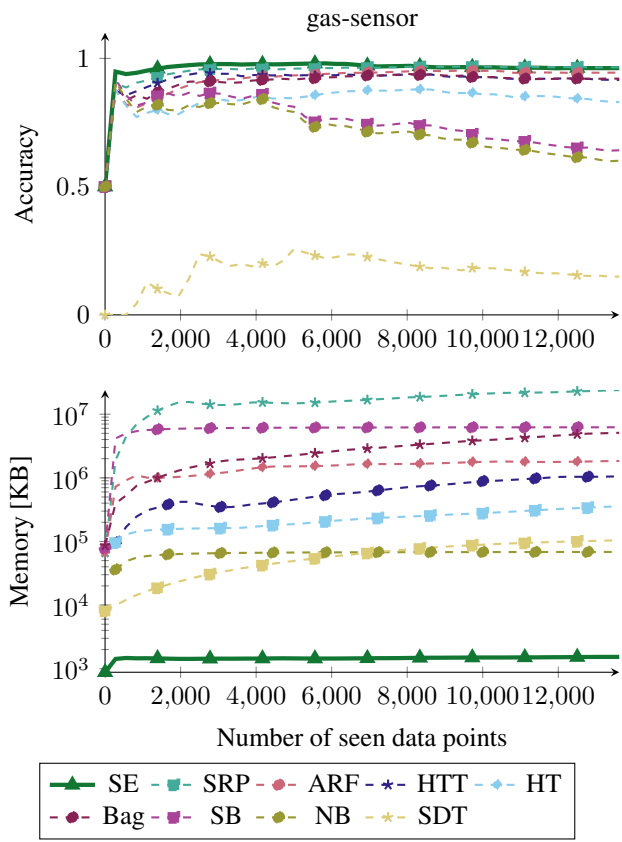

\centering
\input{filecontents/acc_gas}
\input{filecontents/mem_gas}

\pgfplotsset{
SEStyle/.style={col1,mark=triangle*,mark options ={col1},mark repeat={5},ultra thick, error bars/.cd,y dir = both, y explicit},
SRPStyle/.style={col8,dashed,mark options ={col8},mark repeat={5}, thick, error bars/.cd,y dir = both, y explicit},
ARFStyle/.style={col2,dashed,mark options ={col2},mark repeat={5}, thick, error bars/.cd,y dir = both, y explicit},
HTTStyle/.style={col5, dashed,mark options ={col5},mark repeat={5}, thick, error bars/.cd,y dir = both, y explicit},
HTStyle/.style={col4,dashed,mark options ={col4},mark repeat={5}, thick, error bars/.cd,y dir = both, y explicit},
SBStyle/.style={col3,dashed,mark options ={col3},mark repeat={5}, thick, error bars/.cd,y dir = both, y explicit},
NBStyle/.style={col6,thick,dashed,mark options ={col6},mark repeat={5}, thick, error bars/.cd,y dir = both, y explicit},
SDTStyle/.style={col7,thick,dashed,mark options ={col7},mark repeat={5}, thick, error bars/.cd,y dir = both, y explicit},
BagStyle/.style={col9,thick,dashed,mark options ={col9},mark repeat={5}, thick, error bars/.cd,y dir = both, y explicit},
xticklabel style={
        /pgf/number format/fixed,
},
scaled x ticks=false
}

\begin{tikzpicture}
    \begin{groupplot}[group style={group size= 1 by 2, vertical sep=1cm},
    	height=.3\textwidth,
    	width=\columnwidth,
        legend columns =5,
        axis lines = left,
        ]
        \nextgroupplot[ylabel={Accuracy},
        	title={gas-sensor},
        	title style={yshift=-2ex},
        	ymin=0.0, ymax=1.1,
        	legend to name=zelda]
        	\addplot+[SEStyle]  table[x=x,y=y] {accuracy_gas-sensor_SE.dat};
        	\addlegendentry{SE};
        	\addplot+[SRPStyle]  table[x=x,y=y] {accuracy_gas-sensor_SRP.dat};
            \addlegendentry{SRP};
            \addplot+[ARFStyle]  table[x=x,y=y] {accuracy_gas-sensor_ARF.dat};
            \addlegendentry{ARF};
            \addplot+[HTTStyle]  table[x=x,y=y] {accuracy_gas-sensor_HTT.dat};
            \addlegendentry{HTT};
            \addplot+[HTStyle]  table[x=x,y=y] {accuracy_gas-sensor_HT.dat};
            \addlegendentry{HT};
            \addplot+[BagStyle]  table[x=x,y=y] {accuracy_gas-sensor_Bag.dat};
            \addlegendentry{Bag};
        	\addplot+[SBStyle]  table[x=x,y=y] {accuracy_gas-sensor_SB.dat};
            \addlegendentry{SB};
        	\addplot+[NBStyle]  table[x=x,y=y] {accuracy_gas-sensor_NB.dat};
            \addlegendentry{NB};
            \addplot+[SDTStyle]  table[x=x,y=y] {accuracy_gas-sensor_SDT.dat};
            \addlegendentry{SDT};
        \nextgroupplot[
            ylabel={Memory [KB]},
            xlabel=Number of seen data points,
        	xmin=0, ymin=0,
        	ymode=log]
        	\addplot+[NBStyle]  table[x=x,y=y] {memory_gas-sensor_NB.dat};
        	\addplot+[SDTStyle]  table[x=x,y=y] {memory_gas-sensor_SDT.dat};
            \addplot+[SEStyle]  table[x=x,y=y] {memory_gas-sensor_SE.dat};
            \addplot+[SRPStyle]  table[x=x,y=y] {memory_gas-sensor_SRP.dat};
            \addplot+[ARFStyle]  table[x=x,y=y] {memory_gas-sensor_ARF.dat};
            \addplot+[HTTStyle]  table[x=x,y=y] {memory_gas-sensor_HTT.dat};
            \addplot+[HTStyle]  table[x=x,y=y] {memory_gas-sensor_HT.dat};
            \addplot+[SBStyle]  table[x=x,y=y] {memory_gas-sensor_SB.dat};
            \addplot+[BagStyle]  table[x=x,y=y] {memory_gas-sensor_Bag.dat};
    \end{groupplot} 
\end{tikzpicture}\\

\pgfplotslegendfromname{zelda}
\caption{Accuracy and memory consumption over the number of items on the gas-sensor dataset. Best viewed in color.}
\label{fig:gas-sensor}
\end{figure}

\begin{figure}[t!]
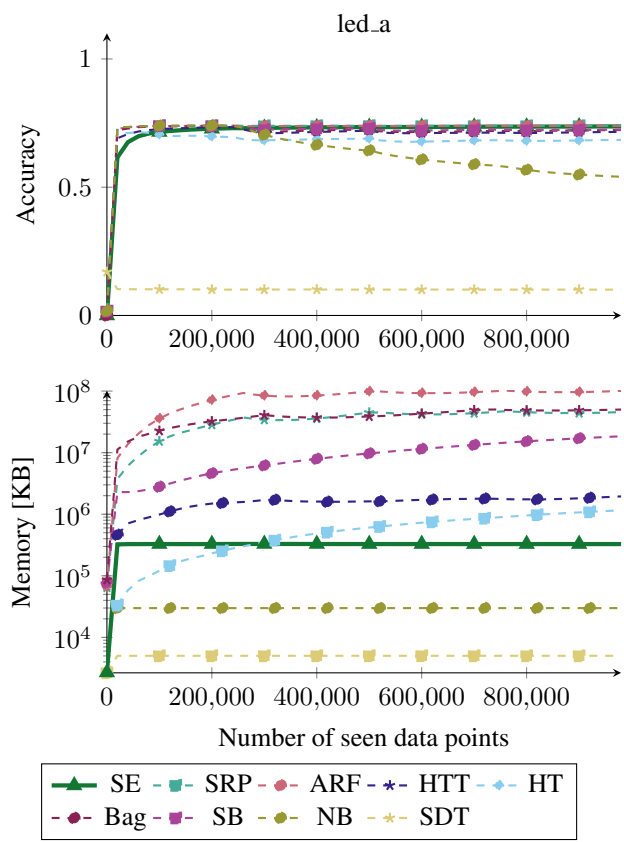

\centering
\input{filecontents/acc_led}
\input{filecontents/mem_led}

\pgfplotsset{
SEStyle/.style={col1,mark=triangle*,mark options ={col1},mark repeat={5},ultra thick, error bars/.cd,y dir = both, y explicit},
SRPStyle/.style={col8,dashed,mark options ={col8},mark repeat={5}, thick, error bars/.cd,y dir = both, y explicit},
ARFStyle/.style={col2,dashed,mark options ={col2},mark repeat={5}, thick, error bars/.cd,y dir = both, y explicit},
HTTStyle/.style={col5, dashed,mark options ={col5},mark repeat={5}, thick, error bars/.cd,y dir = both, y explicit},
HTStyle/.style={col4,dashed,mark options ={col4},mark repeat={5}, thick, error bars/.cd,y dir = both, y explicit},
SBStyle/.style={col3,dashed,mark options ={col3},mark repeat={5}, thick, error bars/.cd,y dir = both, y explicit},
NBStyle/.style={col6,thick,dashed,mark options ={col6},mark repeat={5}, thick, error bars/.cd,y dir = both, y explicit},
SDTStyle/.style={col7,thick,dashed,mark options ={col7},mark repeat={5}, thick, error bars/.cd,y dir = both, y explicit},
BagStyle/.style={col9,thick,dashed,mark options ={col9},mark repeat={5}, thick, error bars/.cd,y dir = both, y explicit},
xticklabel style={
        /pgf/number format/fixed,
},
scaled x ticks=false
}

\begin{tikzpicture}
    \begin{groupplot}[group style={group size= 1 by 2, vertical sep=1cm},
    	height=.3\textwidth,
    	width=\columnwidth,
        legend columns =5,
        axis lines = left,
        ]
        \nextgroupplot[ylabel={Accuracy},
        	title={led\_a},
        	title style={yshift=-2ex},
        	ymin=0.0, ymax=1.1,
        	legend to name=zelda]
        	\addplot+[SEStyle]  table[x=x,y=y] {accuracy_led_a_SE.dat};
        	\addlegendentry{SE};
        	\addplot+[SRPStyle]  table[x=x,y=y] {accuracy_led_a_SRP.dat};
            \addlegendentry{SRP};
            \addplot+[ARFStyle]  table[x=x,y=y] {accuracy_led_a_ARF.dat};
            \addlegendentry{ARF};
            \addplot+[HTTStyle]  table[x=x,y=y] {accuracy_led_a_HTT.dat};
            \addlegendentry{HTT};
            \addplot+[HTStyle]  table[x=x,y=y] {accuracy_led_a_HT.dat};
            \addlegendentry{HT};
            \addplot+[BagStyle]  table[x=x,y=y] {accuracy_led_a_Bag.dat};
            \addlegendentry{Bag};
        	\addplot+[SBStyle]  table[x=x,y=y] {accuracy_led_a_SB.dat};
            \addlegendentry{SB};
        	\addplot+[NBStyle]  table[x=x,y=y] {accuracy_led_a_NB.dat};
            \addlegendentry{NB};
            \addplot+[SDTStyle]  table[x=x,y=y] {accuracy_led_a_SDT.dat};
            \addlegendentry{SDT};
        \nextgroupplot[
            ylabel={Memory [KB]},
            xlabel={Number of seen data points},
        	ymode=log,
        	]
        	\addplot+[NBStyle]  table[x=x,y=y] {memory_led_a_NB.dat};
        	\addplot+[SDTStyle]  table[x=x,y=y] {memory_led_a_SDT.dat};
            \addplot+[SEStyle]  table[x=x,y=y] {memory_led_a_SE.dat};
            \addplot+[SRPStyle]  table[x=x,y=y] {memory_led_a_SRP.dat};
            \addplot+[ARFStyle]  table[x=x,y=y] {memory_led_a_ARF.dat};
            \addplot+[HTTStyle]  table[x=x,y=y] {memory_led_a_HTT.dat};
            \addplot+[HTStyle]  table[x=x,y=y] {memory_led_a_HT.dat};
            \addplot+[SBStyle]  table[x=x,y=y] {memory_led_a_SB.dat};
            \addplot+[BagStyle]  table[x=x,y=y] {memory_led_a_Bag.dat};
    \end{groupplot} 
\end{tikzpicture}\\

\pgfplotslegendfromname{zelda}
\caption{Accuracy and memory consumption over the number of items on the led\_a dataset. Best viewed in color.}
\label{fig:led_a}
\end{figure}

To gain a more complete picture we now inspect the iterative development of the test-then-train accuracy and the model size over the learning process for the best performing configuration of each algorithm without any memory constraints. Figures \ref{fig:gas-sensor} and \ref{fig:led_a} plot the number of seen data points against the accuracy and memory requirement for the gas-sensor and led\_a dataset. The gas-sensor dataset is interesting because it contains real-world data with a known time of drift whereas the led\_a dataset contain artificial drift.
Looking at the accuracy in Figure \ref{fig:gas-sensor} (top row) we notice a rather chaotic behavior in the beginning, which can be attributed to the fact that in the first 10 months of measurements new classes appear for the first time. Once each class was presented at-least once to the algorithms, the accuracy approaches one, where SE has the highest accuracy, followed by SRP. After roughly $4,000$ datapoints we see a drop in the accuracy of ARF and NB which can be attributed to sudden changes in the distribution after roughly 20 month of measurements. Here, the number of measurements, as well as the class distribution heavily changes. We also observe that HTT seems to cope better with the dynamic changes in this dataset, compared to HT, which can be expected from the more greedy nature of the algorithm. 

Looking at the memory consumption in Figure \ref{fig:gas-sensor} (bottom row) we notice an interesting behavior (note the logarithmic scale). We see that SE uses by far the fewest, strictly bounded resources whereas the other algorithm require at least a magnitude more memory. As expected, the memory consumption of HT and HTT monotonously rises over time as these algorithms never remove any internal node from the tree. Moreover, HTT must maintain the list of all possible splits at all times, thereby requiring more memory than HT. Likewise, the use of multiple HT(T)s as base learners of ARF, SRP, Bagging and SB is reflected in the plot. In addition, ARF and SRP both utilize the ADWIN drift detector \cite{bifet/gavalda/2007} which uses a variable sized sliding window. The window size is computed by Hoeffding's Bound which is ideally suited if no specific distribution can be assumed. On the downside, the window size convergences comparably slow. As a result, the algorithms store additional information for large windows to detect a possible drift further increasing the memory consumption.
Naive Bayes (NB) is a strong competitor to SE, but also requires roughly a magnitude more memory. Last, we notice that the memory consumption of SDT increases over time due to a large sliding window in the hyperparameter settings.
We conclude that our SE method offers the best predictive performance in the gas-sensor dataset while using the fewest resources making it an ideal algorithm for resource-constrained environments.

Figure \ref{fig:led_a} shows the test-then-train accuracy (top row) and average model size (bottom row) for the led\_a dataset. 
While the accuracy is relatively stable in the beginning we can see a clear drop around the 250,000 item mark and a smaller drop later at around 500,000 items. NB and HT seem to suffer the most from this concept drift, but also the other algorithms lose some predictive power. Again, SDT does not seem to learn anything at-all. Looking at the memory consumption, we see a similar picture as before: HT(T) and ensembles of HT(T) learners steadily increase their memory consumption over time, requiring up to 100 MB. NB and SDT have the smallest memory consumption wheres SE ranks third in memory consumption in this setting. Again, we find that SE offers excellent performance while ranking among the most resource-friendly methods.

\section{Conclusions} 
\label{sec:conclusion}

In this paper we introduced Shrub Ensembles (SE) as a tree ensemble which is able to process large amounts of data with limited resources in a fast-paced way, while adapting to new situations on the fly. We noticed, that incremental tree learner never remove any nodes from the tree slowly increasing the memory consumption over time. Gradient-based tree learning, on the other hand, requires the costly computation of gradients through backpropagation, making the overall algorithm slow. 
In contrast, Shrub Ensembles are ensembles of small- to medium sized trees which are aggressively pruned during optimization, such that sub-optimal shrubs are removed while new shrubs are regularly introduced. We theoretically showed that shrub ensembles will always add a new tree to the ensemble once a new, previously unknown concept is available (cf.\@ Theorem~\ref{th:learning-shrubs}). Further, we discussed the behavior of our algorithm on noisy data and the influence of its hyperparameters. Our Shrub Ensembles retain an excellent performance even when only little memory is available. In an extensive experimental study we showed that SE offers a better accuracy-memory trade-off in 7 of 12 cases, while having a statistically significant better performance than most other methods.

\section*{Acknowledgments}
Part of the work on this paper has been supported by Deutsche Forschungsgemeinschaft (DFG) within the Collaborative Research Center SFB 876 "Providing Information by Resource-Constrained Analysis", DFG project number 124020371, SFB project A1, \url{http://sfb876.tu-dortmund.de}. Part of the work on this research has been funded by the Federal Ministry of Education and Research of Germany as part of the competence center for machine learning ML2R (01|18038A), \url{https://www.ml2r.de/}. 

\bibliography{literatur}
\end{document}


\title{APPENDIX: Shrub Ensembles for Online Classification}

\author{\name Sebastian Buschjäger \email sebastian.buschjaeger@tu-dortmund.de\\
\name Sibylle Hess \email s.c.hess@tue.nl\\
\name Katharina Morik \email katharina.morik@tu-dortmund.de}

\maketitle
 
\begin{abstract}
This appendix accompanies the paper `Shrub Ensembles for Online Classification'. It provides results for more experiments which are not given in the paper due to space reasons. 
\end{abstract}

\section{Dataset}

Table \ref{tab:datasets} gives an overview of the datasets used for all experiments. All datasets are freely available online. The detailed download scripts for each dataset are provided in the anonymized version of the source code.

\begin{table}[H]
\centering
\caption{\label{tab:datasets} Characteristics of employed datasets. The top group are real-world datasets with unknown concept drift and the bottom group are artificial datasets with synthetic drift.}
\begin{tabular}{@{}lrrrc@{}}
\toprule
Dataset     & N     & d   & C \\ \midrule
gas-sensor & 13~910 & 128 & 5     \\
weather  & 18~159 & 8 & 2  \\
nomao  & 34~465 & 174 &  2       \\
elec  & 45~312 &  14 & 2  \\
airlines & 539~383 & 614 & 2     \\ 
covertype  &581~012& 98 & 7\vspace{0.2cm} \\
agrawal\_a & 1~000~000 & 40 & 2 \\ 
agrawal\_g & 1~000~000 & 40 & 2 \\ 
led\_a & 1~000~000 & 48 & 10   \\ 
led\_g & 1~000~000 & 48 & 10   \\ 
rbf\_f & 1~000~000 & 10  & 5   \\ 
rbf\_m & 1~000~000 & 10  & 5   \\ \bottomrule
\end{tabular}
\end{table}

Figure \ref{fig:config} gives an overview of the hyper-parameter optimization involved. For each method a dictionary of hyperparameter has been defined. The \texttt{Var} keyword marks a variation of the corresponding hyperparameter which randomly selects one of the values from the list. For example, the depicted configuration might generate a SE model with $M=128$ trees, trained on a window-size of $2^{10}$ data points with a step size of $0.1$ where each tree has a max depth of $12$ which is randomly trained on $d$ features. The detailed list of hyperparameters can be found in the source code. 

\begin{figure}[H]
\caption{Example of a hyperparameter configuration for Shrub Ensembles (SE). The \texttt{Var} keyword marks a variation of the corresponding hyperparameter which randomly selects one of the values from the list. All trees receive the same hyperparameter configuration. Please consult the source code for the exact parameter choices for all methods.} 
\label{fig:config}
\begin{center}
    \begin{minipage}{0.5\textwidth}
        \begin{minted}[fontsize=\footnotesize]{python}
        "M": Var([4,8,16,32,64,128,256]),
        "window_size": Var([2**i for i in range(4, 14)]),
        "step_size": Var([1e-4,1e-3,1e-2,1e-1,2e-1,5e-1]),
        "additional_tree_options" : {
            "max_depth": Var([2,4,8,12,15]),
            "splitter": Var(["train","random"]),
            "max_features": Var([d, np.sqrt(d)])
        }
        \end{minted}
    \end{minipage}
\end{center}
\end{figure}

\section{Proof of Theorem \ref{th:learning-shrubs}}

\begin{theorem}
\label{th:learning-shrubs}
Let $M \ge 1$ be the maximum ensemble size in the shrub ensembles (SE) algorithm and let $B$ be the buffer size.
Consider a classification problem with $c$ classes.
Further, let $m \le M$ be the number of models in the ensemble. Now assume that a new observation $(x_B,y_B)$ arrives, which was previously unknown to the ensemble so that $\forall i=1,\dots,m \colon h_i(x_B) \not= y_{B}$. Let SE train fully-grown trees with $h_j(x)\in \{0,1\}^C$ and let $h$ be the new tree, trained on the current window, such that $\forall i=1,\dots,B \colon h(x_i) = y_{i}$. Then we have for $\alpha > \frac{BC}{4m}$ the following cases:

\begin{itemize}
    \item (1) If $m < M$, then $h$ is added to the ensemble
    \item (2) If $m = M$, then $h$ replaces the tree with the smallest weight from the ensemble.
\end{itemize}
\end{theorem}

\begin{proof}

We start by computing the gradient for weight $w$ with the corresponding tree $h$:

\begin{align*}
    \frac{\partial \ell(f(x), y)}{\partial w} &= \frac{2}{BC} \left( \sum_{i=1}^{B} \sum_{c=1}^C (f(x_i)_c - y_{i,c}) h(x_i)_c  \right) \\
\end{align*}

Now consider the first case $m < M$. We show, that $h$ receives a non-negative weight $w > 0$ and thus is kept after applying the prox-operator. To do so, we check the weight of $h$ after the gradient step. Recall, that new trees receive an initial weight of $w = 0$ and thus:
$$
w = 0 - \frac{2\alpha}{BC} \left( \sum_{i=1}^{B} \sum_{c=1}^C (f(x_i)_c - y_{i,c}) h(x_i)_c \right) \overset{?}{>} 0
$$
simplifying and reordering leads to
\begin{align*}
\sum_{i=1}^{B} \sum_{c=1}^C (f(x_i)_c - y_{i,c}) h(x_i)_c &\overset{?}{<} 0 \\
\sum_{i=1}^{B} \sum_{c=1}^C f(x_i)_c h(x_i)_c &\overset{?}{<} \sum_{i=1}^{B} \sum_{c=1}^C h(x_i)_c y_{i,c}
\end{align*}

Note that $h$ is a fully grown tree on the current batch so that $h(x_i)_c$ and $y_{i,c}$ are either both $0$ or both $1$. Thus, it follows that since only one entry in $y_i$ is nonzero that the right side equals $B$. Now consider the left side. Recall that $h(x_B)_c$ is $1$ for the new class $j$ and $f(x_B)_j = 0$ per assumption. Thus, in the `best' case the ensembles prediction $f$ and the prediction of $h$ is the same on all but the last example. More formally, $\sum_{c=1}^C f(x_i)_c h(x_i)_c = 1$ for each, but the last item in $\mathcal B$. It follows that:
$$
\sum_{i=1}^{B} \sum_{c=1}^C f(x_i)_c h(x_i)_c = B -1 < B 
$$
which concludes the proof for the first case.

Now we consider the second case in which $m=M$. Here we must show that after the gradient step the weight $w$ is larger than the smallest weight in the ensemble to replace the corresponding tree. Let $w_k$ be the smallest weight in the ensemble. As noted, $h$ replaces that tree $h_k$ with the smallest weight $w_k$ in the ensemble if $w_k < w$ after the gradient step. Let
$$
G_k = \frac{2}{BC}\sum_{i=1}^{B-1} \sum_{c=1}^C (f(x_i)_c - y_{i,c})h_k(x_i)_c 
$$ be the gradient on the first $B-1$ examples in the window for the $k-$th member. Now consider the extreme case in which $h_k$ is always correct but $f$ is always wrong. Then $G_k$ is at-least
$$
G_k = \frac{2}{BC}\sum_{i=1}^{B-1} \sum_{c=1}^C f(x_i)_c h_k(x_i)_c - y_{i,c}h_k(x_i)_c \ge -\frac{2 (B-1)}{BC}
$$
A similar argument holds for $h$, since we know per assumption that $h$ is correct on the entire batch of the training data. Since $\sum_{i=1}^m w_i = 1$ we can furthere estimate that $w_k \le \frac{1}{m}$. Thus:
\begin{align*}
    w_k - &\alpha G_k - \frac{\alpha}{BC} \sum_{c=1}^C 2(f(x_B)_c - y_{B,c})h_k(x)_c \\
    &\quad\quad < w_k + \frac{2(B-1)}{BC} - \frac{\alpha}{BC} \sum_{c=1}^C 2(f(x_B)_c - y_{B,c})h_k(x)_c \\
    &\quad\quad < \frac{1}{m} + \frac{2(B-1)}{BC} - \frac{\alpha}{BC} \sum_{c=1}^C 2(f(x_B)_c - y_{B,c})h_k(x)_c \\
    &\quad\quad \overset{?}{<} w = 0 + \frac{2(B-1)}{BC} - \frac{\alpha}{BC} \sum_{c=1}^C 2(f(x_B)_c - y_{B,c})h(y)_c
\end{align*}
Subtracting $\frac{2(B-1)}{BC}$ on both sides leads to:
\begin{align*}
\frac{1}{m} - \frac{2\alpha}{BC} &\sum_{c=1}^C (f(x_B)_c - y_{B,c})h_k(x)_c \\
&\quad\quad \overset{?}{<} - \frac{2\alpha}{BC} \sum_{c=1}^C (f(x_B)_c - y_{B,c})h(x)_c
\end{align*}
Looking at the left side we note that $f(x_B)_j$ and $h_k(x)_j$ are $0$ for the class of $y_B$ and $h_k(x)_c = 1$ for exactly one other class $c$. Since $f(x_B)_c \le 1$ we upper bound:
$$
\frac{1}{m} - \frac{2\alpha}{BC}\sum_{c\not= j} f(x_B)_c h_k(x_B)_c < \frac{1}{m} - \frac{2\alpha}{BC}
$$
Looking at the right side we note that $h(x)_j = 1$ whereas the remaining entries are $0$. Moreover, $f(x_B)_j = 0$ and $y_{B,c} = 1$ leads to
$$
- \frac{\alpha}{BC} \sum_{c=1}^C 2(f(x_B)_c - y_{B,c})h(x)_c = - \frac{2\alpha}{BC} (-1)
$$
Combining both results:

$$
\frac{1}{m} - \frac{2\alpha}{BC} < \frac{2\alpha}{BC} \Leftrightarrow \alpha > \frac{BC}{4m}
$$
which concludes the proof.




\end{proof}

\section{Additional results}

In addition to the results presented in the paper, we present the raw test-then-train accuracies for the different datasets in this section. To do so, we first select the best performing model without any constraints. After that we apply different constraints to the maximum model size. 

\subsection{Unlimited memory available}
\begin{table}
\centering
\caption{\label{tab:experiments_0} Best test-then-train accuracy for each method on the given datasets without any memory constraints.}

\resizebox{\columnwidth}{!}{%
\begin{tabular}{@{}llrrrrrrrrr@{}}
\toprule
dataset    &               & ARF             & Bag             & HT        & HTT       & NB       & SB         & SDT      & SE (ours)            & SRP             \\ \midrule
gas-sensor & accuracy      & 92.674          & 91.608          & 84.798    & 92.535    & 72.898   & 76.023     & 17.164   & \textbf{96.787} & 95.715          \\
          & size {[}kb{]} & 1796        & 5040        & 355   & 1041  & 67   & 6090   & 103  & \textit{1}           & 23650       \\
          & time {[}s{]}  & 34          & 30          & 14    & 17    & 14   & 112    & 1968 & \textit{8}           & 134\vspace{0.05cm}\\
nomao      & accuracy      & 98.566          & 97.760          & 94.733    & 96.871    & 92.792   & 96.889     & 88.628   & \textbf{98.660} & 98.558          \\
          & size {[}kb{]} & 9292        & 90717       & 1091  & 411   & 42   & 12159  & \textit{7}    & 10          & 15997       \\
          & time {[}s{]}  & 148         & 637         & 26    & 28    & 28   & 147    & 97   & \textit{21}          & 204\vspace{0.05cm}\\
elec       & accuracy      & 91.057          & 89.774          & 86.452    & 87.748    & 76.165   & 88.372     & 42.514   & \textbf{94.012} & 91.415          \\
          & size {[}kb{]} & 5727        & 10640       & 254   & 234   & 3    & 2113   & 4    & \textit{1}           & 18882       \\
          & time {[}s{]}  & 117         & 72          & 17    & 21    & 25   & 56     & 119  & \textit{12}          & 258\vspace{0.05cm}\\
weather    & accuracy      & 78.328          & 78.380          & 75.112    & 76.592    & 69.584   & 76.954     & 32.870   & 75.860          & \textbf{79.192} \\
          & size {[}kb{]} & 22868       & 6034        & 96    & 95    & \textit{2}    & 794    & 106  & 333         & 4012        \\
          & time {[}s{]}  & 95          & 36          & \textit{8}     & 9     & 10   & 23     & 910  & 379         & 27\vspace{0.05cm} \\ 
covtype    & accuracy      & 93.647          & 92.256          & 86.460    & 88.688    & 64.591   & 85.236     & 3.742    & 92.844          & \textbf{94.297} \\
          & size {[}kb{]} & 34204       & 1058285     & 11876 & 10432 & 45   & 51600  & \textit{10}   & 23          & 59589       \\
          & time {[}s{]}  & 2805        & 19558       & \textit{502}   & 611   & 589  & 2150   & 4583 & 1043        & 4527\vspace{0.05cm}\\
airlines   & accuracy      & 69.450          & \textbf{69.501} & 67.004    & 68.756    & 66.969   & 69.867     & 38.771   & 68.176          & 69.067          \\
          & size {[}kb{]} & 203557      & 314285      & 49390 & 16826 & 113  & 348796 & \textit{60}   & 91          & 96996       \\
          & time {[}s{]}  & 5899        & 6801        & 801   & 1038  & \textit{780}  & 5202   & 3041 & 1413        & 3818\vspace{0.05cm}\\
agrawal\_a & accuracy      & \textbf{94.413} & 93.592          & 92.439    & 91.716    & 74.288   & 93.455     & 9.771    & 93.551          & 94.316          \\
          & size {[}kb{]} & 102889      & 216892      & 5909  & 1537  & 8    & 45545  & \textit{5}    & 73          & 121708      \\
          & time {[}s{]}  & 2066        & 2430        & \textit{426}   & 495   & 567  & 1767   & 2782 & 803         & 2741\vspace{0.05cm}\\
agrawal\_g & accuracy      & \textbf{92.277} & 90.192          & 87.589    & 87.707    & 74.276   & 91.189     & 45.478   & 91.567          & 91.894          \\
          & size {[}kb{]} & 132295      & 370692      & 10809 & 2807  & 8    & 61971  & \textit{5}    & 79          & 194785      \\
          & time {[}s{]}  & 2382        & 3176        & \textit{454}   & 497   & 563  & 1876   & 2601 & 827         & 3313\vspace{0.05cm}\\
led\_a     & accuracy      & 73.789          & 72.277          & 68.849    & 71.480    & 64.417   & 72.754     & 10.112   & 72.329          & \textbf{73.855} \\
          & size {[}kb{]} & 99900       & 49406       & 1154  & 1936  & 29   & 18363  & \textit{5}    & 320         & 45368       \\
          & time {[}s{]}  & 4528        & 3452        & \textit{1061}  & 1201  & 1192 & 4361   & 4567 & 1628        & 3256\vspace{0.05cm}\\
led\_g     & accuracy      & 73.036          & 72.114          & 68.634    & 71.406    & 64.401   & 72.495     & 0    & 71.440          & \textbf{73.109} \\
          & size {[}kb{]} & 115453      & 83044       & 1146  & 2407  & 29   & 18429  & \textit{5}    & 324         & 184343      \\
          & time {[}s{]}  & 5401        & 5128        & \textit{1089}  & 1220  & 1231 & 3727   & 4716 & 1602        & 6713\vspace{0.05cm}\\
rbf\_f     & accuracy      & 75.788          & 56.659          & 38.186    & 40.326    & 29.747   & 42.529     & 7.246    & 73.028          & \textbf{77.928} \\
          & size {[}kb{]} & 11285       & 629872      & 1644  & 4599  & 5    & 15575  & \textit{4}    & 241         & 35487       \\
          & time {[}s{]}  & 3109        & 9081        & \textit{542}   & 585   & 708  & 1469   & 4519 & 1271        & 7079\vspace{0.05cm}\\
rbf\_m     & accuracy      & \textbf{86.376} & 81.335          & 62.963    & 69.513    & 32.948   & 65.467     & 16.672   & 79.632          & 86.134          \\
          & size {[}kb{]} & 28217       & 589415      & 2726  & 5013  & 5    & 18409  & \textit{2}    & 916         & 73123       \\
          & time {[}s{]}  & 3251        & 5754        & \textit{547}   & 609   & 710  & 1537   & 2653 & 6468        & 7754        \\\bottomrule
\end{tabular}
}
\end{table}
In our first experiment, we assume that there are no memory restrictions at all. Table \ref{tab:experiments_0} displays the test-then-train accuracy, the average model size (in KB) and the total runtime (in seconds). The highest accuracies are highlighted and the fastest runtime and smallest memory consumption are marked in italic.
We notice that we obtain in comparison to \cite{Gomes/etal/2019} slightly better results for all algorithms, which can be attributed to our more expensive hyperparameter optimization. SE, SRP, ARF, and Bagging generally offer the best accuracy. SRP offers the highest accuracy in 5 cases, followed by SE and ARF, performing best on 3 datasets. Bagging attains the highest accuracy on the airlines dataset. Looking at the size, however, we see a different picture. Most of the well-performing methods in terms of accuracy (SRP, ARF and Bagging) generally consume the most memory, ranging in the order of megabytes, up to the hundreds. SDT, HT and HTT place themselves in the middle, consuming a few hundred kilobytes to a few megabytes, whereas SE and NB generally consume the least memory ($\leq1$ MB). Likewise, looking at the running time, we observe that SRP, ARF, SB and Bagging require the most time, although SDT shows here some notable outliers. HT, HTT, and NB seem to be the quickest methods wheres SE sometimes is very fast and sometimes is slower than HT or HTT. We conclude that SE is among the state-of-the-art algorithms for online learning, while having a similar resource consumption as Online Naive Bayes, making it an ideal choice for small devices.

\subsection{Limited memory available}
In the second experiment we assume that we are given a limited memory budget and we exclude every method which exceeds this limit. Table \ref{tab:experiments_1} shows the results when a maximum of $10$ MB per model is allowed and Table \ref{tab:experiments_2} shows the results if only $1$ MB is available. The best method is again highlighted for each dataset. A dash `-' indicates that an algorithm did not produce a model with a size in the given bounds for any hyperparameter configuration. For space reasons we exclude the specific model size and runtime from the tables.

\begin{table}[t!]
\setlength{\tabcolsep}{2pt}
\centering
\caption{\label{tab:experiments_1} Best test-then-train accuracy for each method on the given datasets with model size below $10$ MB. The best method is depicted in bold for each dataset. A dash `-' indicates that a method did not have a model with a smaller size than the constraint in any hyperparameter configuration.}
\resizebox{\columnwidth}{!}{%
\begin{tabular}{@{}lrrrrrrrrr@{}}
\toprule
dataset    & ARF    & Bag    & HT     & HTT    & NB     & SB     & SDT    & SE~~ & SRP             \\ 
& & & & & & & & (ours) &\\ \midrule
gas-sensor & 92.674          & 91.608 & 84.798 & 92.535          & 72.898 & 76.023 & 17.164 & \textbf{96.787} & 95.617          \\
nomao      & 98.566          & 97.315 & 94.733 & 96.871          & 92.792 & 96.451 & 88.628 & \textbf{98.660} & 98.515          \\
elec       & 91.057          & 89.122 & 86.452 & 87.748          & 76.165 & 88.372 & 42.514 & \textbf{94.012} & 91.020          \\
weather    & 78.244          & 78.380 & 75.112 & 76.592          & 69.584 & 76.954 & 32.870 & 75.860          & \textbf{79.192} \\
covtype    & \textbf{93.194} & -      & 86.408 & 88.505          & 64.591 & 82.922 & 3.742  & 92.844          & 92.958          \\
airlines   & -               & -      & -      & \textbf{68.491} & 66.969 & -      & 38.771 & 68.176          & -               \\
agrawal\_a & -               & 93.029 & 92.439 & 91.716          & 74.288 & 92.879 & 9.771  & \textbf{93.551} & -               \\
agrawal\_g & -               & 87.399 & 87.356 & 87.707          & 74.276 & 88.817 & 45.478 & \textbf{91.567} & -               \\
led\_a     & \textbf{73.501} & 70.518 & 68.849 & 71.480          & 64.417 & 72.331 & 10.112 & 72.329          & 73.042          \\
led\_g     & \textbf{72.680} & 70.266 & 68.634 & 71.406          & 64.401 & 72.090 & -      & 71.440          & 71.764          \\
rbf\_f     & 75.744          & 45.391 & 38.186 & 40.326          & 29.747 & 41.210 & 7.246  & 73.028          & \textbf{77.444} \\
rbf\_m     & \textbf{85.030} & 69.880 & 62.963 & 69.513          & 32.948 & 63.004 & 16.672 & 79.632          & 84.908          \\ \bottomrule
\end{tabular}
}
\end{table}
Looking at Table \ref{tab:experiments_1}, we first notice that SE now also offers the best performance on the agrawal\_a and agrwal\_g dataset, because neither ARF nor SRP managed to produce models which use less than $10$ MB. Also interestingly, HTT suddenly becomes the best method on the airlines dataset and ARF manages to become the best method on the led\_a and led\_g dataset. 

\begin{table}
\setlength{\tabcolsep}{2pt}
\centering
\caption{\label{tab:experiments_2} Best test-then-train accuracy for each method on the given datasets with model size below $1$ MB. The best method is depicted in bold for each dataset. A dash `-' indicates that a method did not have a model with a smaller size than the constraint in any hyperparameter configuration.}
\resizebox{\columnwidth}{!}{%
\begin{tabular}{@{}lrrrrrrrrr@{}}
\toprule
dataset    & ARF    & Bag    & HT     & HTT    & NB     & SB     & SDT    & SE~~ & SRP             \\ 
& & & & & & & & (ours) &\\ \midrule
gas-sensor & 92.336 & 81.173 & 84.798 & 92.177 & 72.898 & 73.471 & 17.164 & \textbf{96.787} & 93.815          \\
nomao      & -      & -      & 94.564 & 96.871 & 92.792 & -      & 88.628 & \textbf{98.660} & -               \\
elec       & 90.453 & 87.839 & 86.452 & 87.748 & 76.165 & 85.965 & 42.514 & \textbf{94.012} & -               \\
weather    & -      & 77.771 & 75.112 & 76.592 & 69.584 & 76.954 & 32.870 & 75.860          & \textbf{77.806} \\
covtype    & -      & -      & -      & -      & 64.591 & -      & 3.742  & \textbf{92.844} & -               \\
airlines   & -      & -      & -      & -      & 66.969 & -      & 38.771 & \textbf{68.176} & -               \\
agrawal\_a & -      & -      & -      & 90.539 & 74.288 & -      & 9.771  & \textbf{93.551} & -               \\
agrawal\_g & -      & -      & -      & 85.846 & 74.276 & -      & 45.478 & \textbf{91.567} & -               \\
led\_a     & -      & -      & -      & -      & 64.417 & -      & 10.112 & \textbf{72.329} & -               \\
led\_g     & -      & -      & -      & -      & 64.401 & -      & -      & \textbf{71.440} & -               \\
rbf\_f     & 69.554 & -      & 36.207 & 35.618 & 29.747 & -      & 7.246  & \textbf{73.028} & -               \\
rbf\_m     & -      & -      & 55.465 & -      & 32.948 & -      & 16.672 & \textbf{79.632} & -               \\ \bottomrule
\end{tabular}
}
\end{table}
We see that this effect amplifies if only $1$ MB is available in Table \ref{tab:experiments_2}. Now, SE is the best method on all datasets except the weather dataset, because ARF, SRP, SB and Bagging do not produce valid models on most datasets. Only SDT, NB and SE manage to consistently stay below $1$ MB. We also performed experiments with more aggressive constraints such as $128$ KB but noticed that most algorithms except NB and SE would fail in this setting while SE generally outperformed NB. We conclude that SE or NB are well-prepared for resource-constraint devices, consistently producing small models. Looking at the accuracy of both approaches we see that SE is the clear winner on all but one dataset. SE is able to produce excellent models while meeting even more aggressive resource constraints making it ideally suited for small, resource-constraint systems.

\subsection{Additional Qualitative Results}

For completeness we also report the test-then-train accuracy and the model size over the number of data points for all dataset. Moreover, we also plot the Pareto front. Please note, that in the paper we re-worked the plots using TIKZ for better visibility. The plots shown here are directly taken exported from matplotlib. 

\begin{figure}[H]
\begin{minipage}{.49\textwidth}
    \centering
    \includegraphics[width=\textwidth,keepaspectratio]{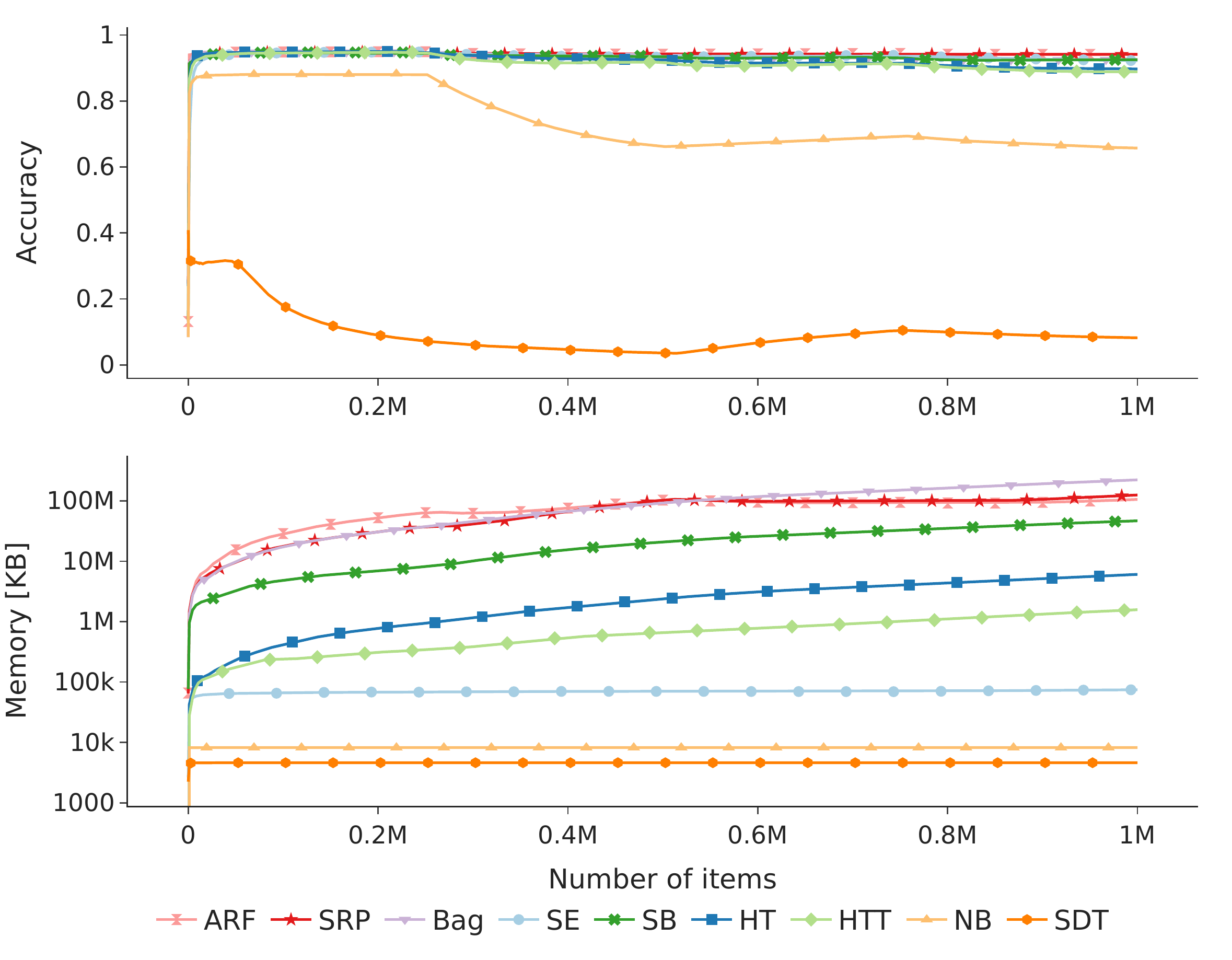}
\end{minipage}\hfill
\begin{minipage}{.49\textwidth}
    \centering 
    \includegraphics[width=\textwidth,keepaspectratio]{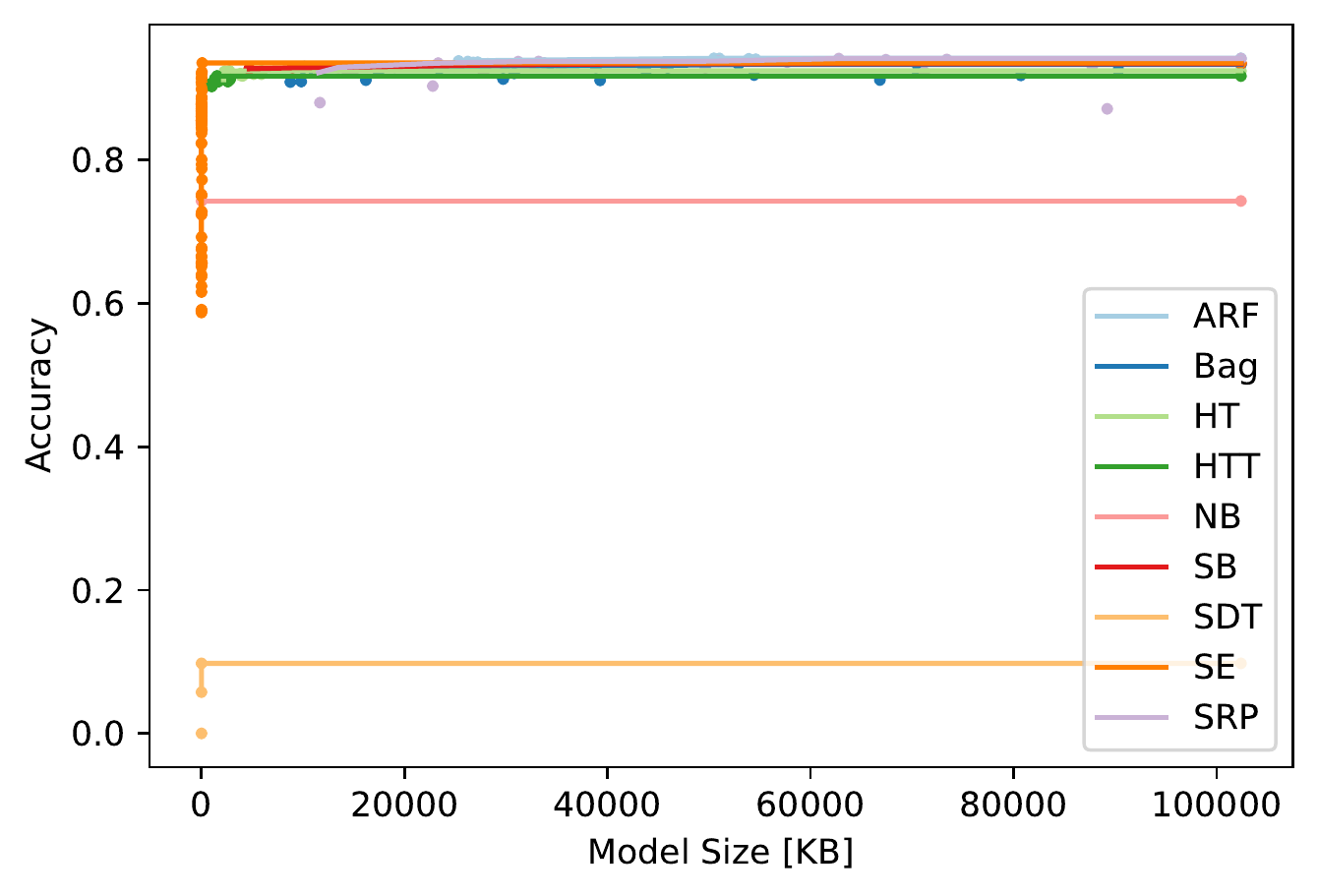}
\end{minipage}
\caption{(left) Test-then-train accuracy and memory consumption on the agrawal\_a dataset of the best configuration over the number of data items in the stream. (right) Pareto front on the agrawal\_a dataset of each method.}
\end{figure}

\begin{figure}[H]
\begin{minipage}{.49\textwidth}
    \centering
    \includegraphics[width=\textwidth,keepaspectratio]{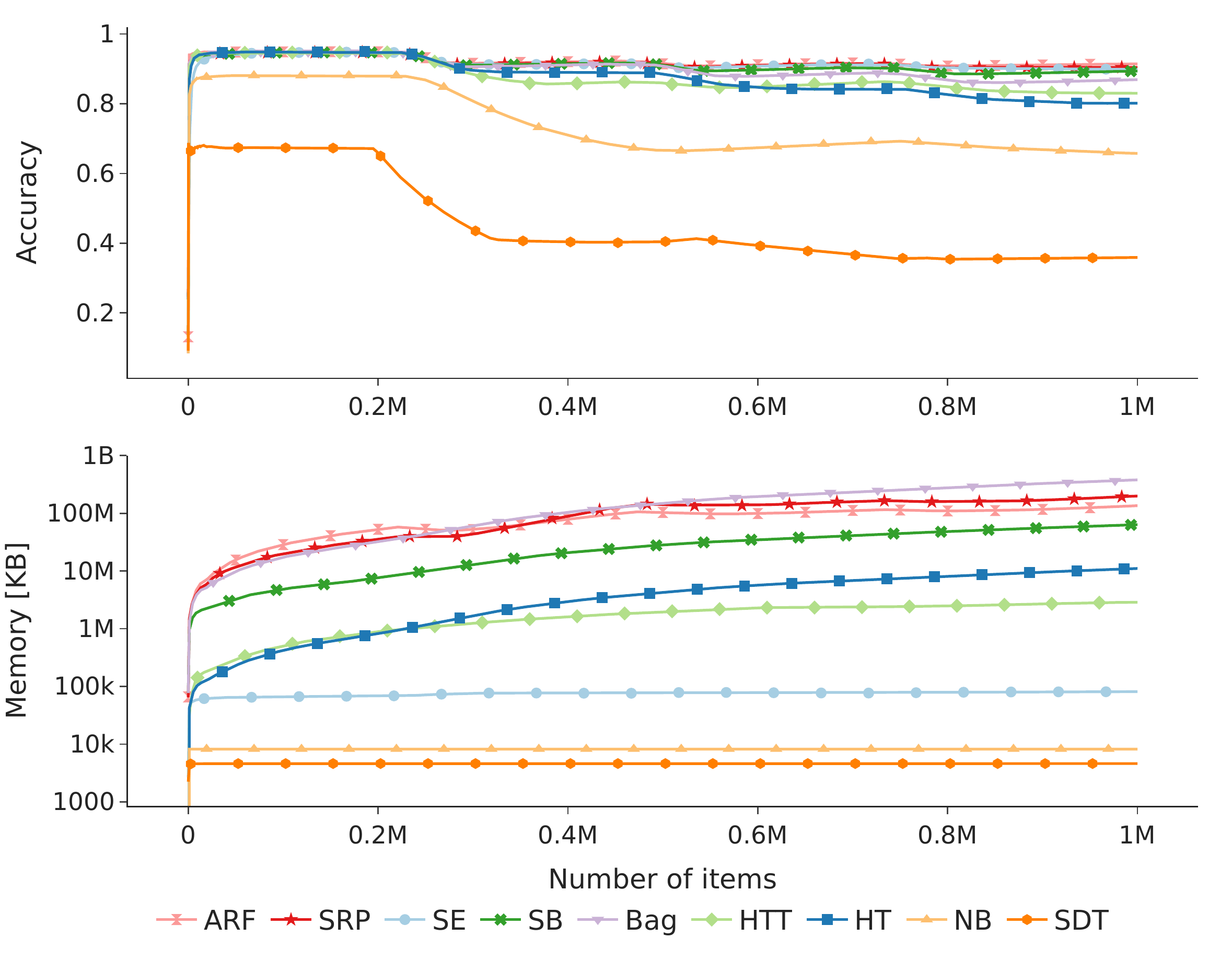}
\end{minipage}\hfill
\begin{minipage}{.49\textwidth}
    \centering 
    \includegraphics[width=\textwidth,keepaspectratio]{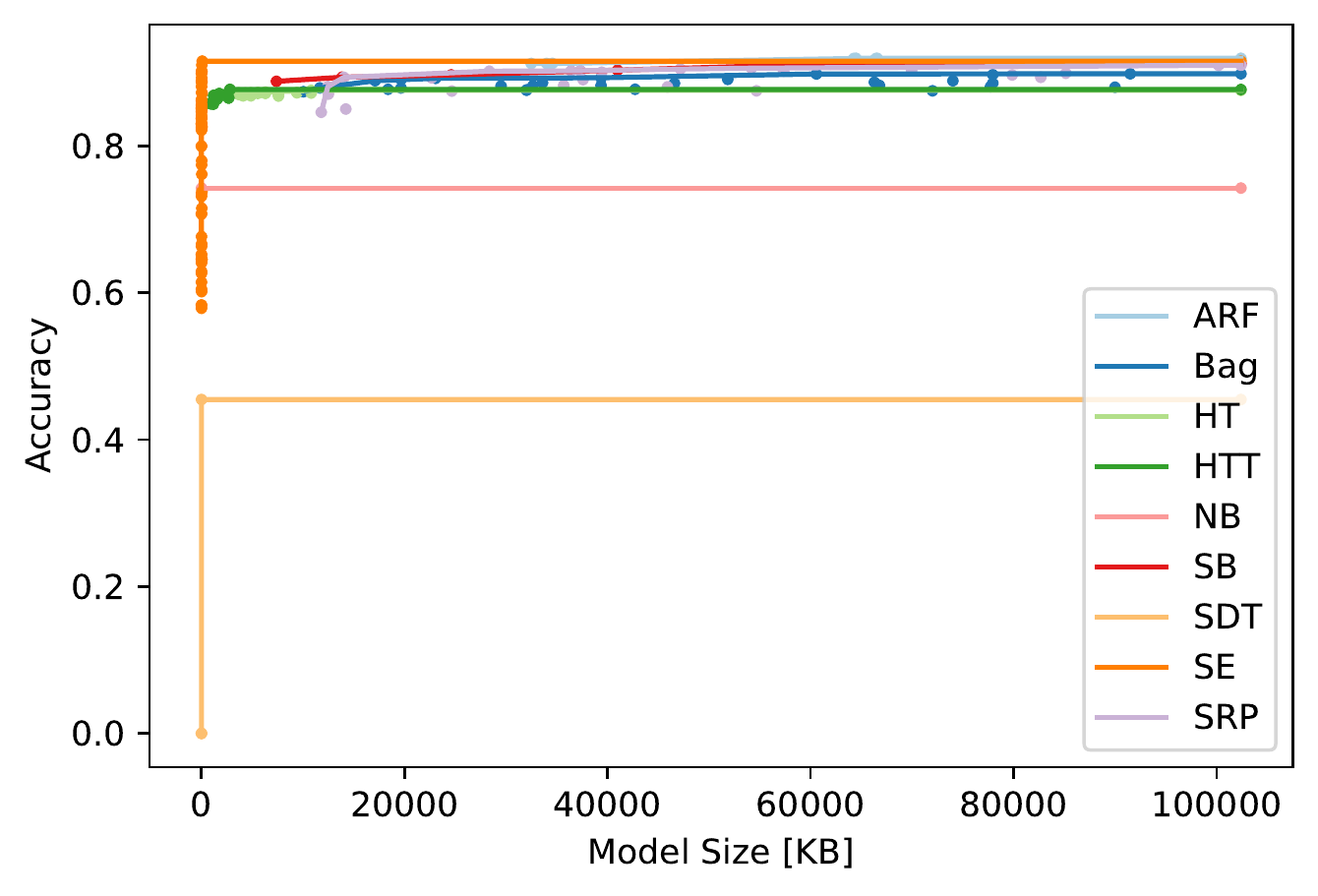}
\end{minipage}
\caption{(left) Test-then-train accuracy and memory consumption on the agrawal\_g dataset of the best configuration over the number of data items in the stream. (right) Pareto front on the agrawal\_g dataset of each method.}
\end{figure}

\begin{figure}[H]
\begin{minipage}{.49\textwidth}
    \centering
    \includegraphics[width=\textwidth,keepaspectratio]{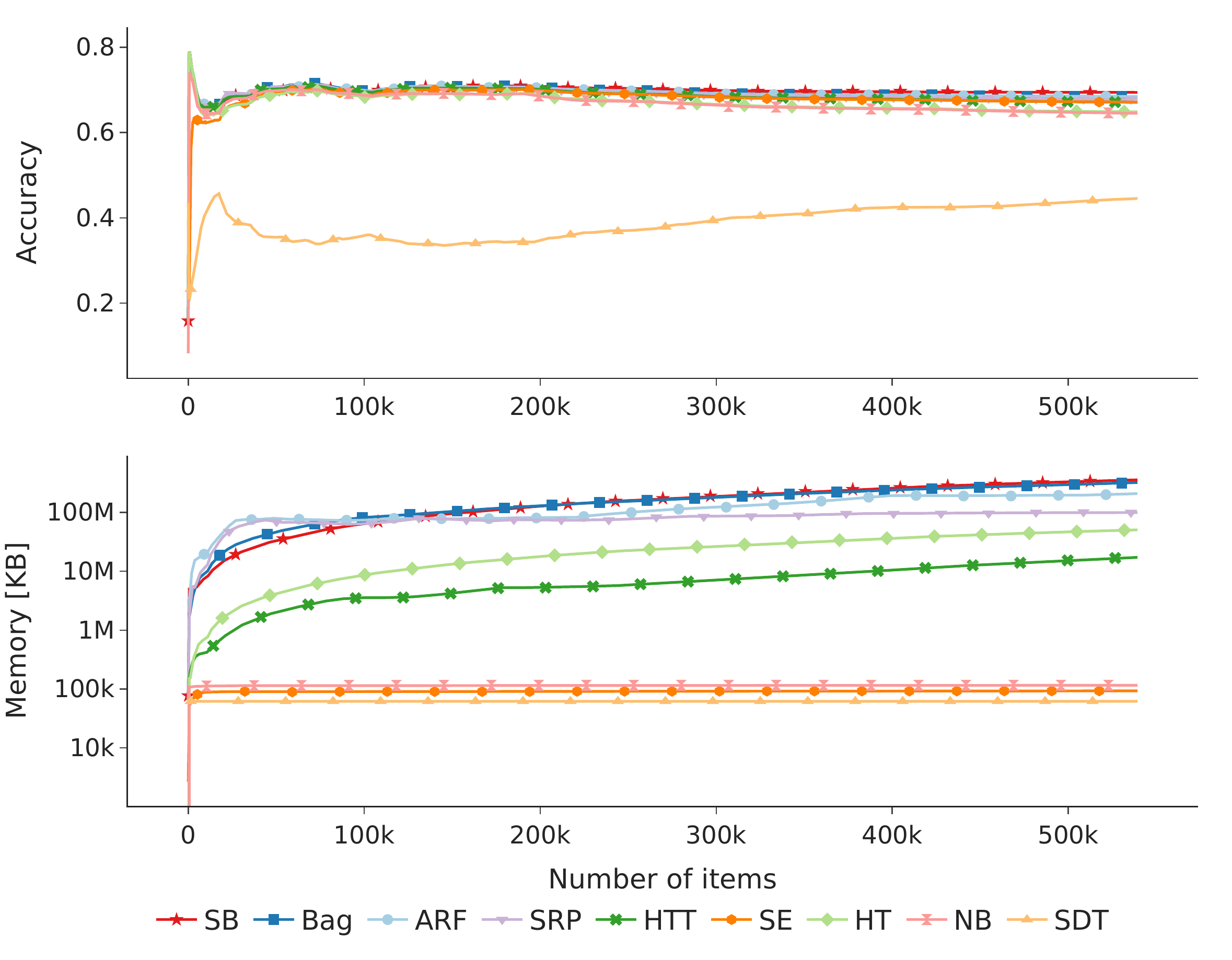}
\end{minipage}\hfill
\begin{minipage}{.49\textwidth}
    \centering 
    \includegraphics[width=\textwidth,keepaspectratio]{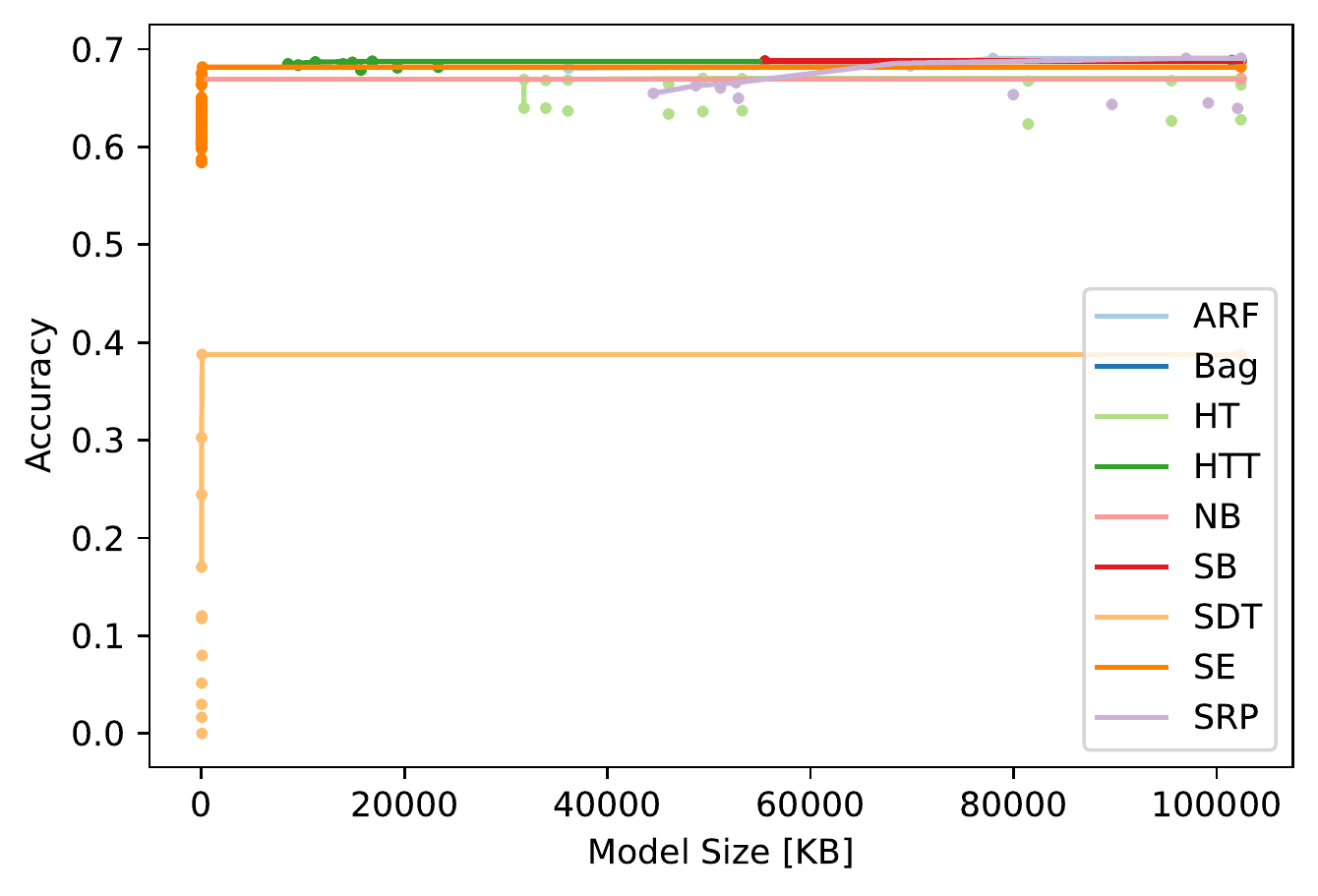}
\end{minipage}
\caption{(left) Test-then-train accuracy and memory consumption on the airlines dataset of the best configuration over the number of data items in the stream. (right) Pareto front on the airlines dataset of each method.}
\end{figure}

\begin{figure}[H]
\begin{minipage}{.49\textwidth}
    \centering
    \includegraphics[width=\textwidth,keepaspectratio]{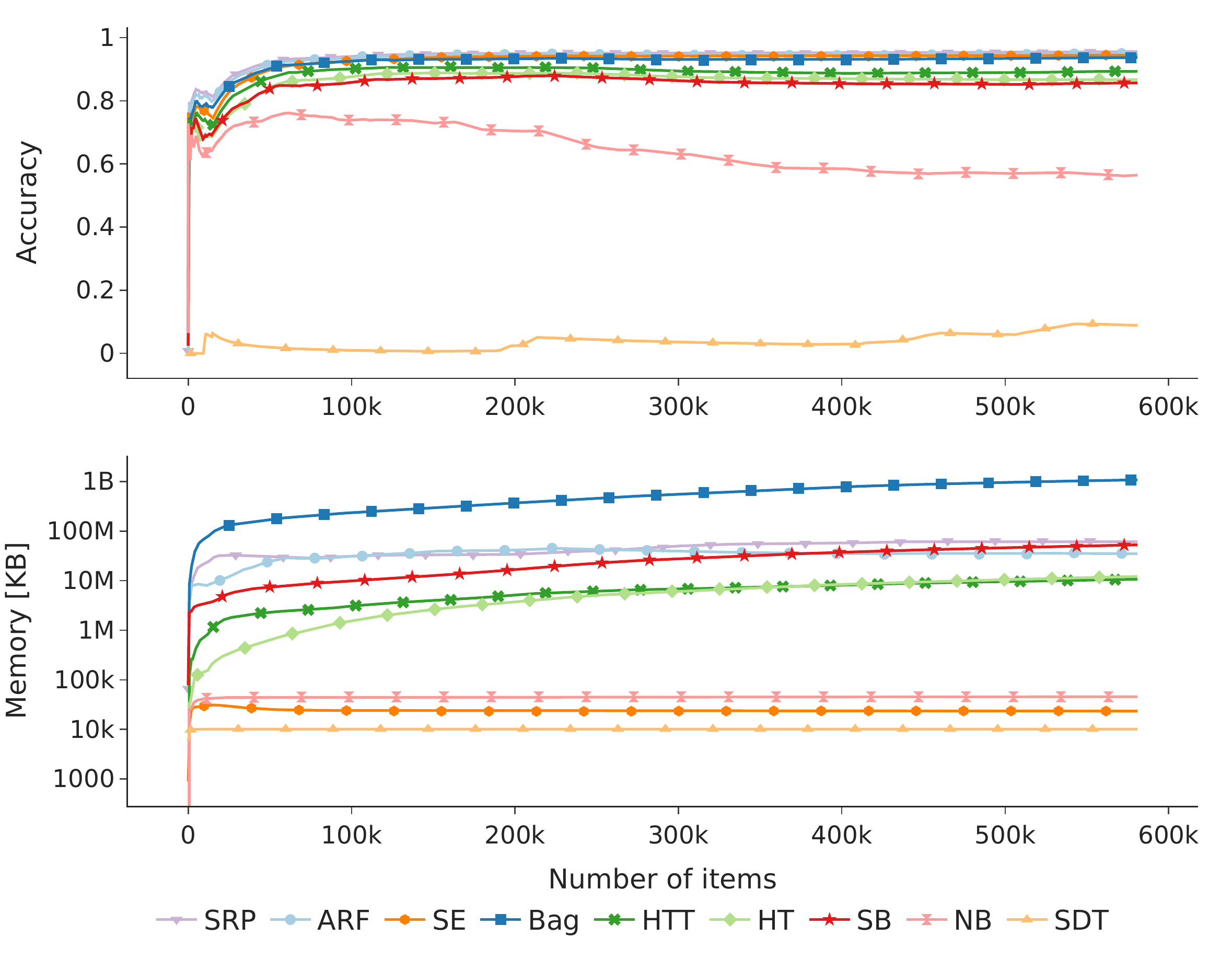}
\end{minipage}\hfill
\begin{minipage}{.49\textwidth}
    \centering 
    \includegraphics[width=\textwidth,keepaspectratio]{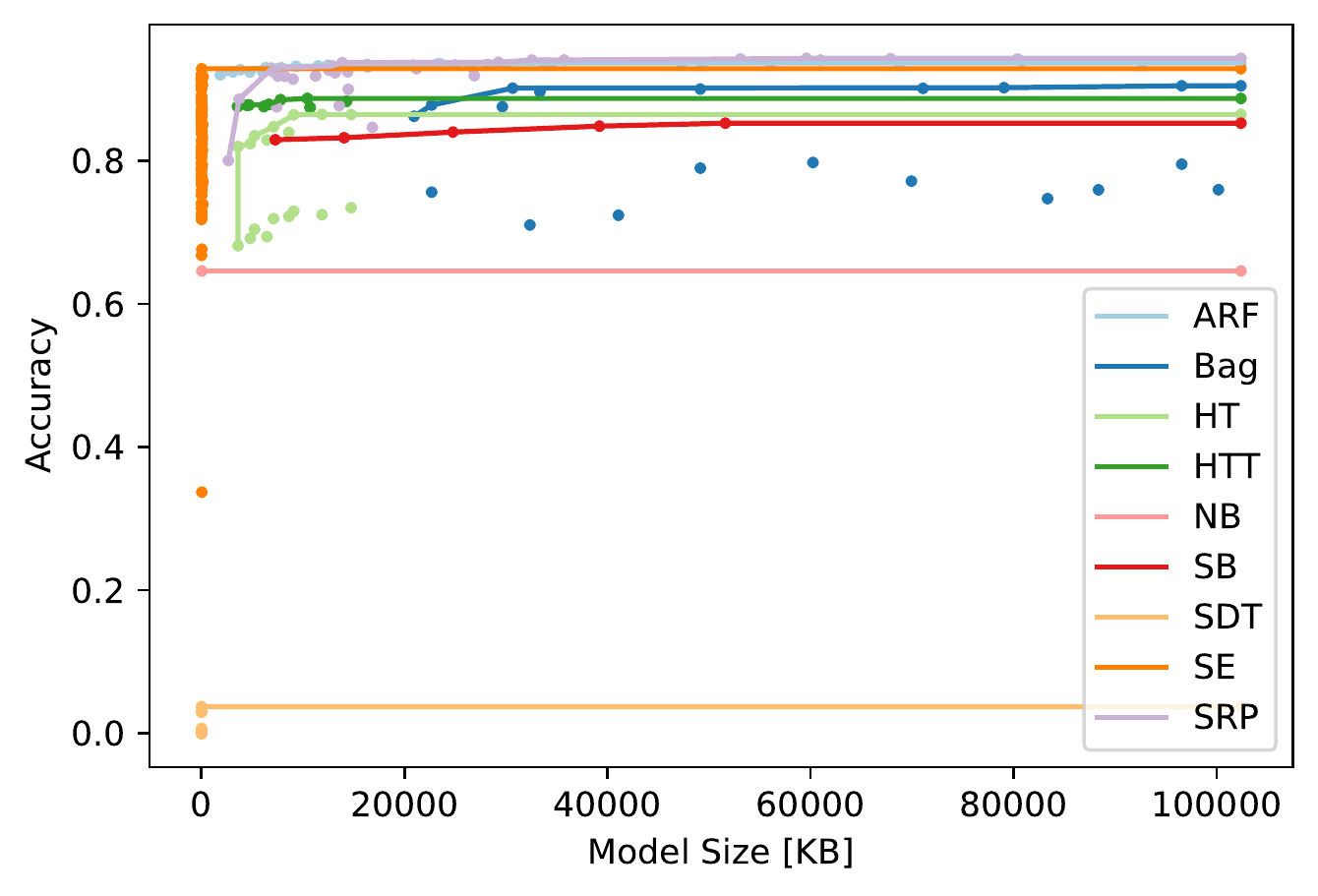}
\end{minipage}
\caption{(left) Test-then-train accuracy and memory consumption on the covtype dataset of the best configuration over the number of data items in the stream. (right) Pareto front on the covtype dataset of each method.}
\end{figure}

\begin{figure}[H]
\begin{minipage}{.49\textwidth}
    \centering
    \includegraphics[width=\textwidth,keepaspectratio]{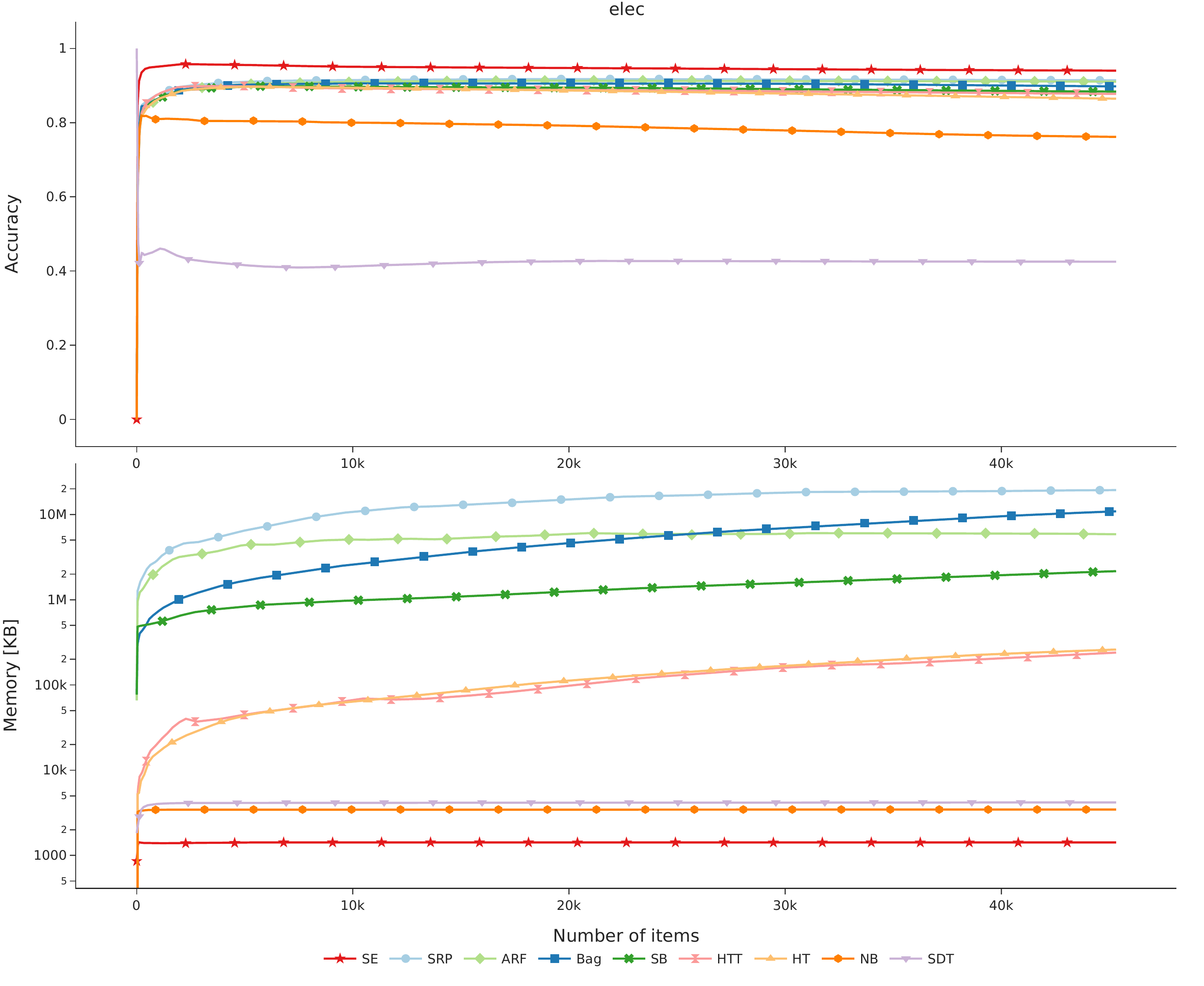}
\end{minipage}\hfill
\begin{minipage}{.49\textwidth}
    \centering 
    \includegraphics[width=\textwidth,keepaspectratio]{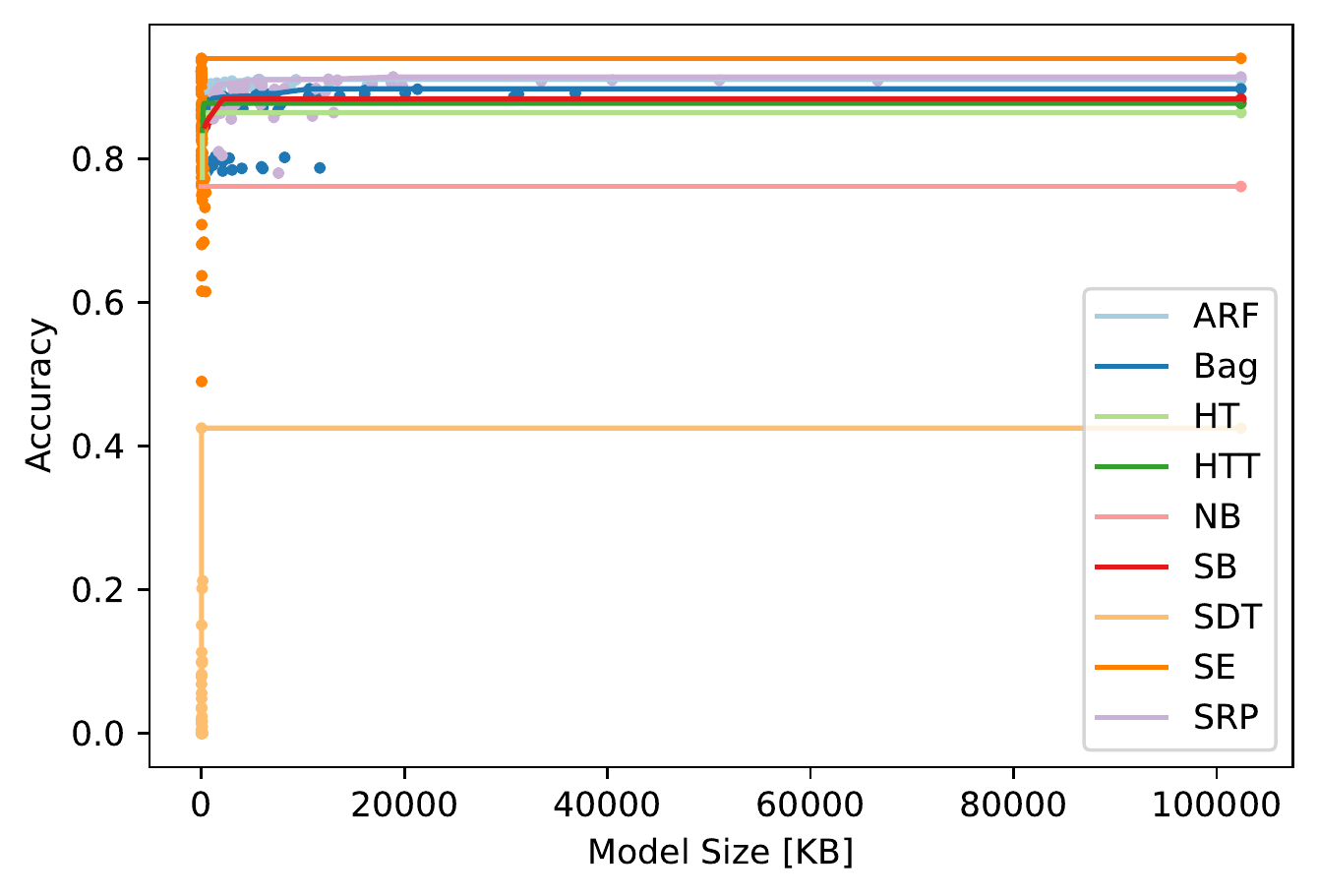}
\end{minipage}
\caption{(left) Test-then-train accuracy and memory consumption on the elec dataset of the best configuration over the number of data items in the stream. (right) Pareto front on the elec dataset of each method.}
\end{figure}

\begin{figure}[H]
\begin{minipage}{.49\textwidth}
    \centering
    \includegraphics[width=\textwidth,keepaspectratio]{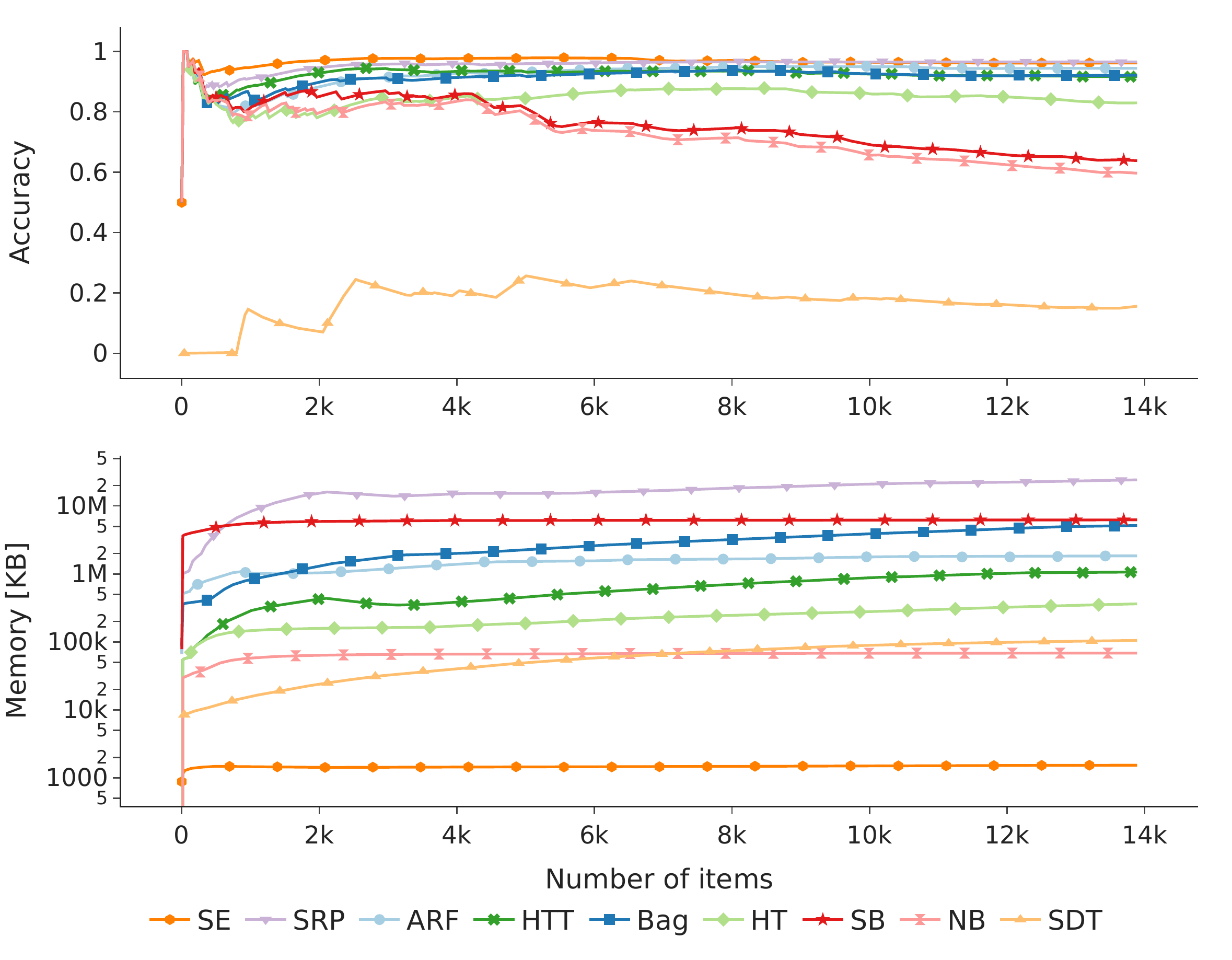}
\end{minipage}\hfill
\begin{minipage}{.49\textwidth}
    \centering 
    \includegraphics[width=\textwidth,keepaspectratio]{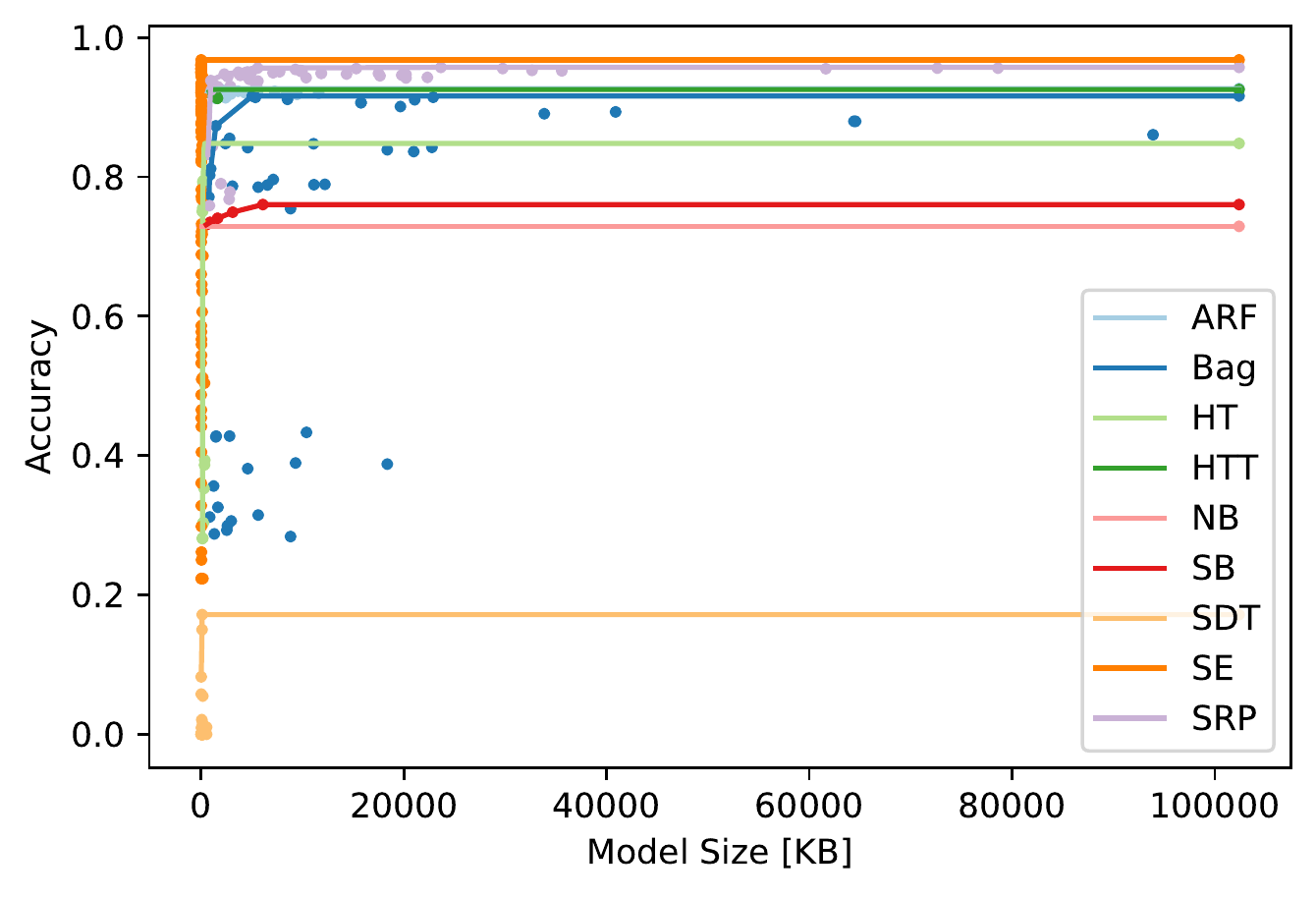}
\end{minipage}
\caption{(left) Test-then-train accuracy and memory consumption on the gas-sensor dataset of the best configuration over the number of data items in the stream. (right) Pareto front on the gas-sensor dataset of each method.}
\end{figure}

\begin{figure}[H]
\begin{minipage}{.49\textwidth}
    \centering
    \includegraphics[width=\textwidth,keepaspectratio]{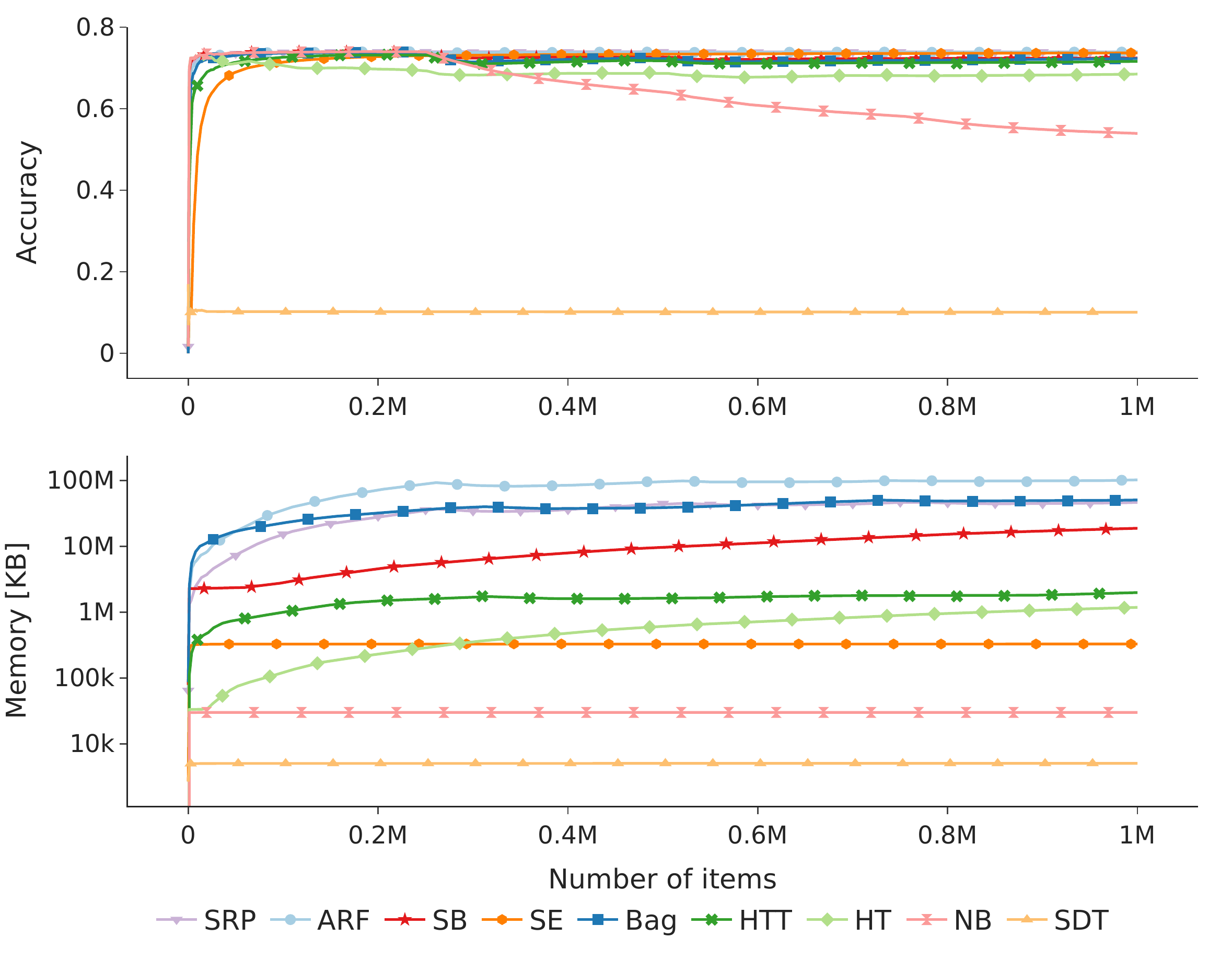}
\end{minipage}\hfill
\begin{minipage}{.49\textwidth}
    \centering 
    \includegraphics[width=\textwidth,keepaspectratio]{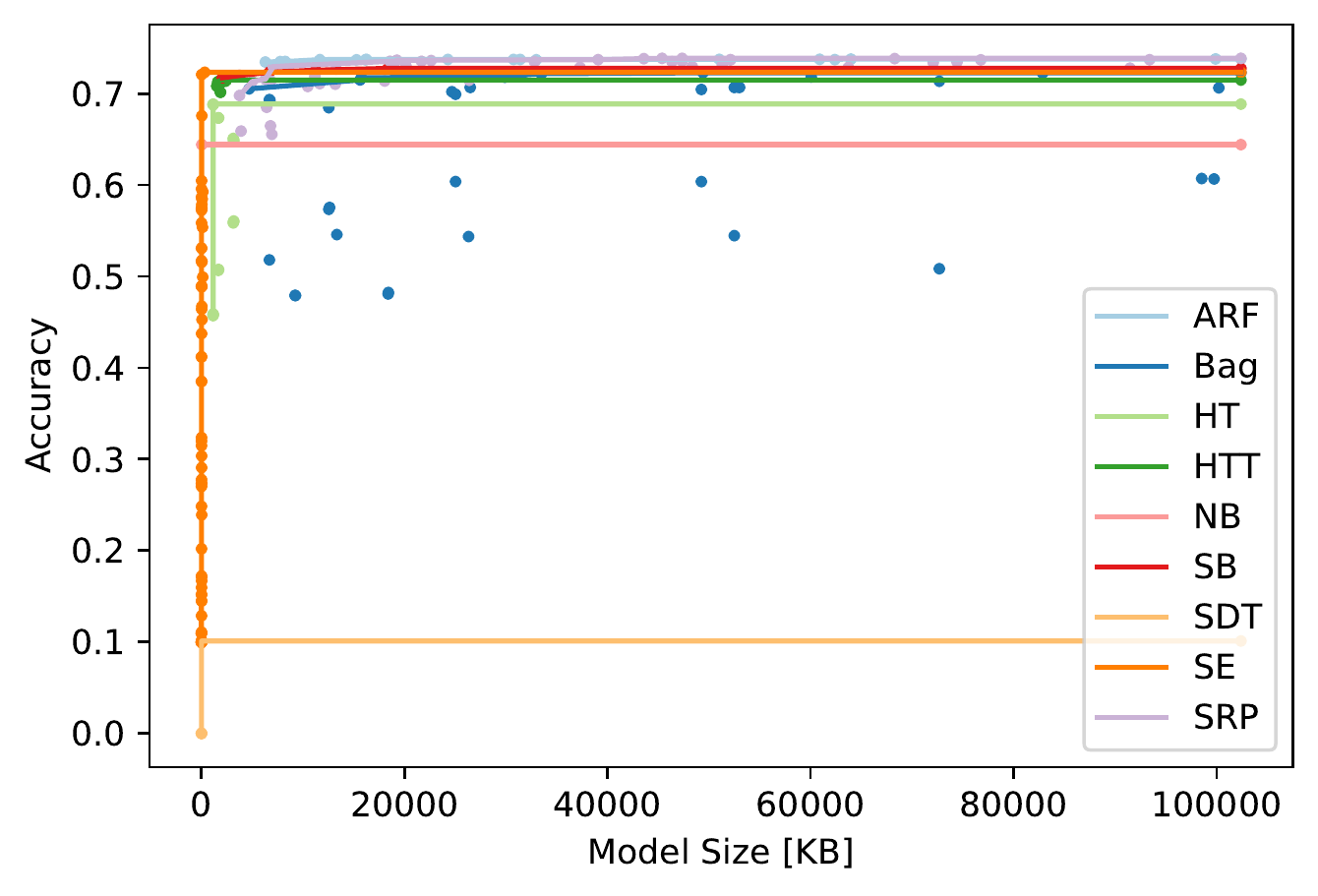}
\end{minipage}
\caption{(left) Test-then-train accuracy and memory consumption on the led\_a dataset of the best configuration over the number of data items in the stream. (right) Pareto front on the led\_a dataset of each method.}
\end{figure}

\begin{figure}[H]
\begin{minipage}{.49\textwidth}
    \centering
    \includegraphics[width=\textwidth,keepaspectratio]{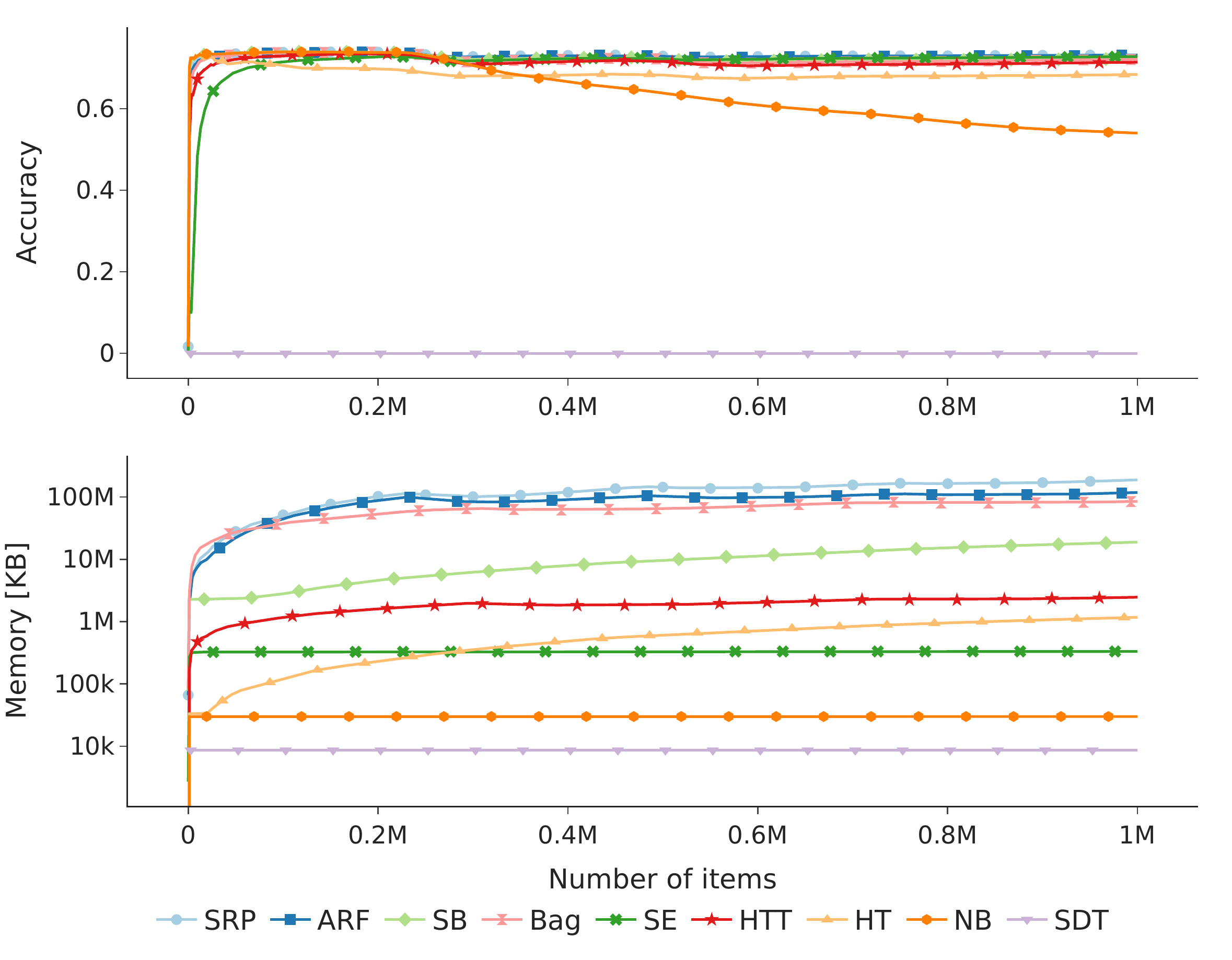}
\end{minipage}\hfill
\begin{minipage}{.49\textwidth}
    \centering 
    \includegraphics[width=\textwidth,keepaspectratio]{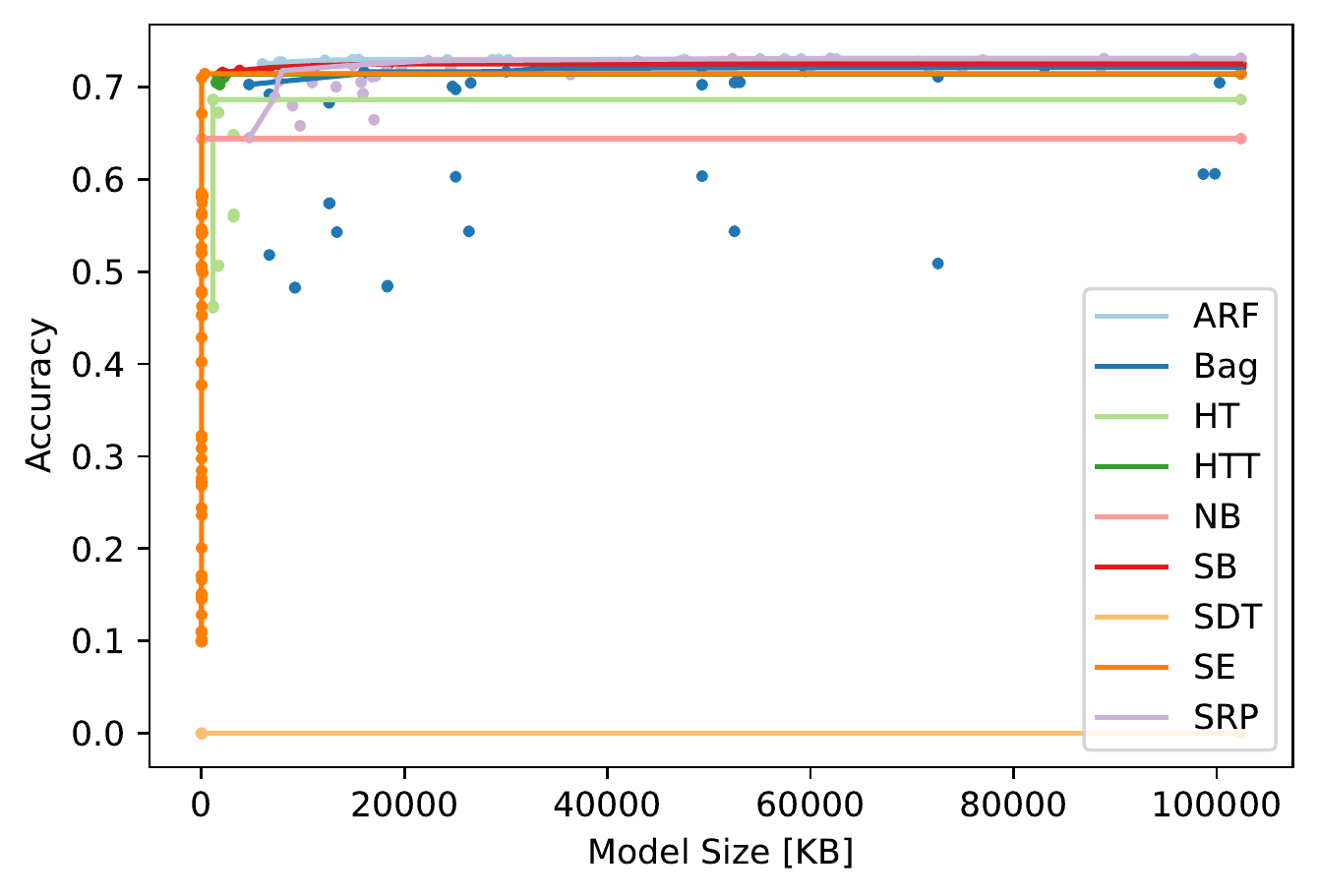}
\end{minipage}
\caption{(left) Test-then-train accuracy and memory consumption on the led\_g dataset of the best configuration over the number of data items in the stream. (right) Pareto front on the led\_g dataset of each method.}
\end{figure}

\begin{figure}[H]
\begin{minipage}{.49\textwidth}
    \centering
    \includegraphics[width=\textwidth,keepaspectratio]{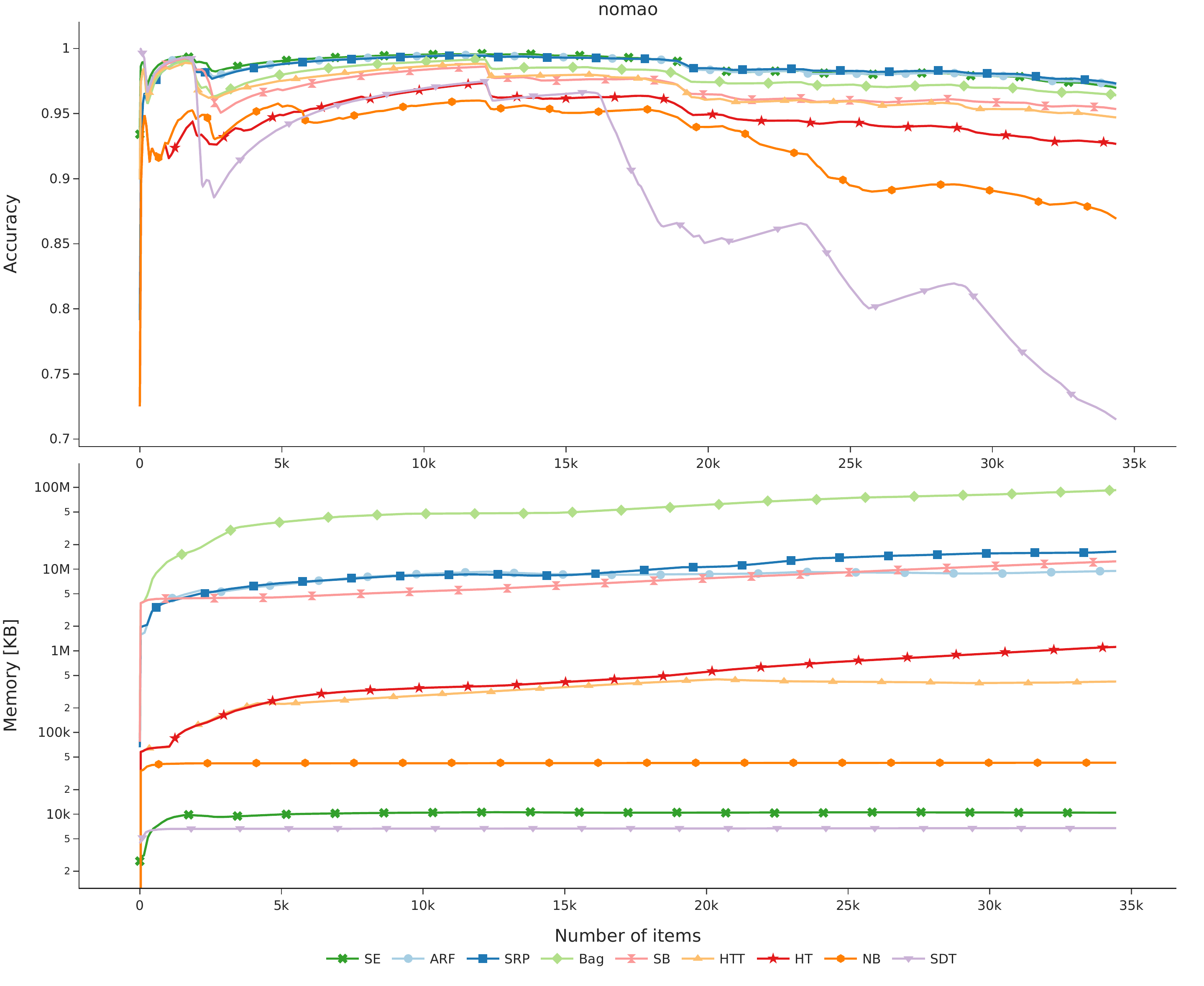}
\end{minipage}\hfill
\begin{minipage}{.49\textwidth}
    \centering 
    \includegraphics[width=\textwidth,keepaspectratio]{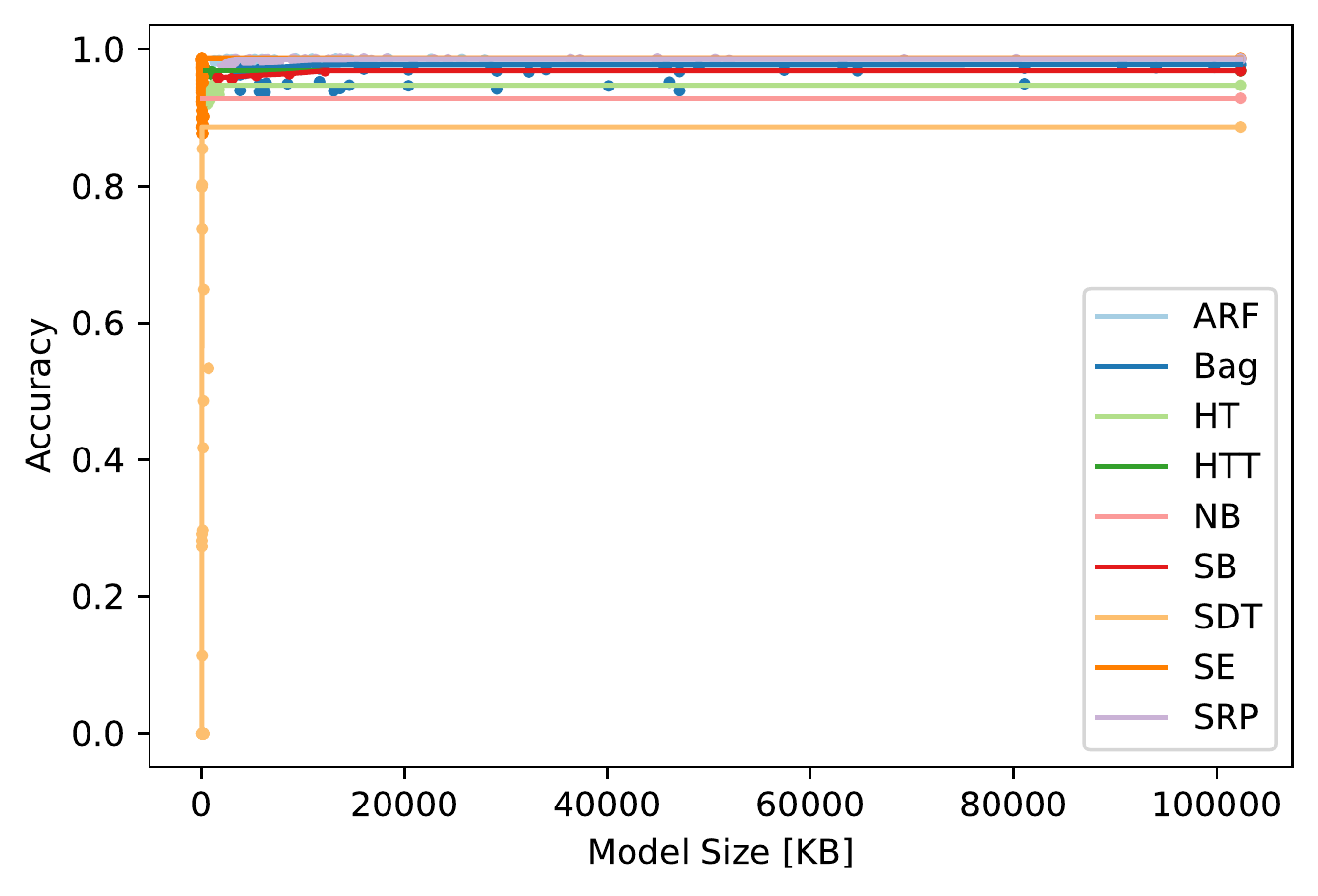}
\end{minipage}
\caption{(left) Test-then-train accuracy and memory consumption on the nomao dataset of the best configuration over the number of data items in the stream. (right) Pareto front on the nomao dataset of each method.}
\end{figure}

\begin{figure}[H]
\begin{minipage}{.49\textwidth}
    \centering
    \includegraphics[width=\textwidth,keepaspectratio]{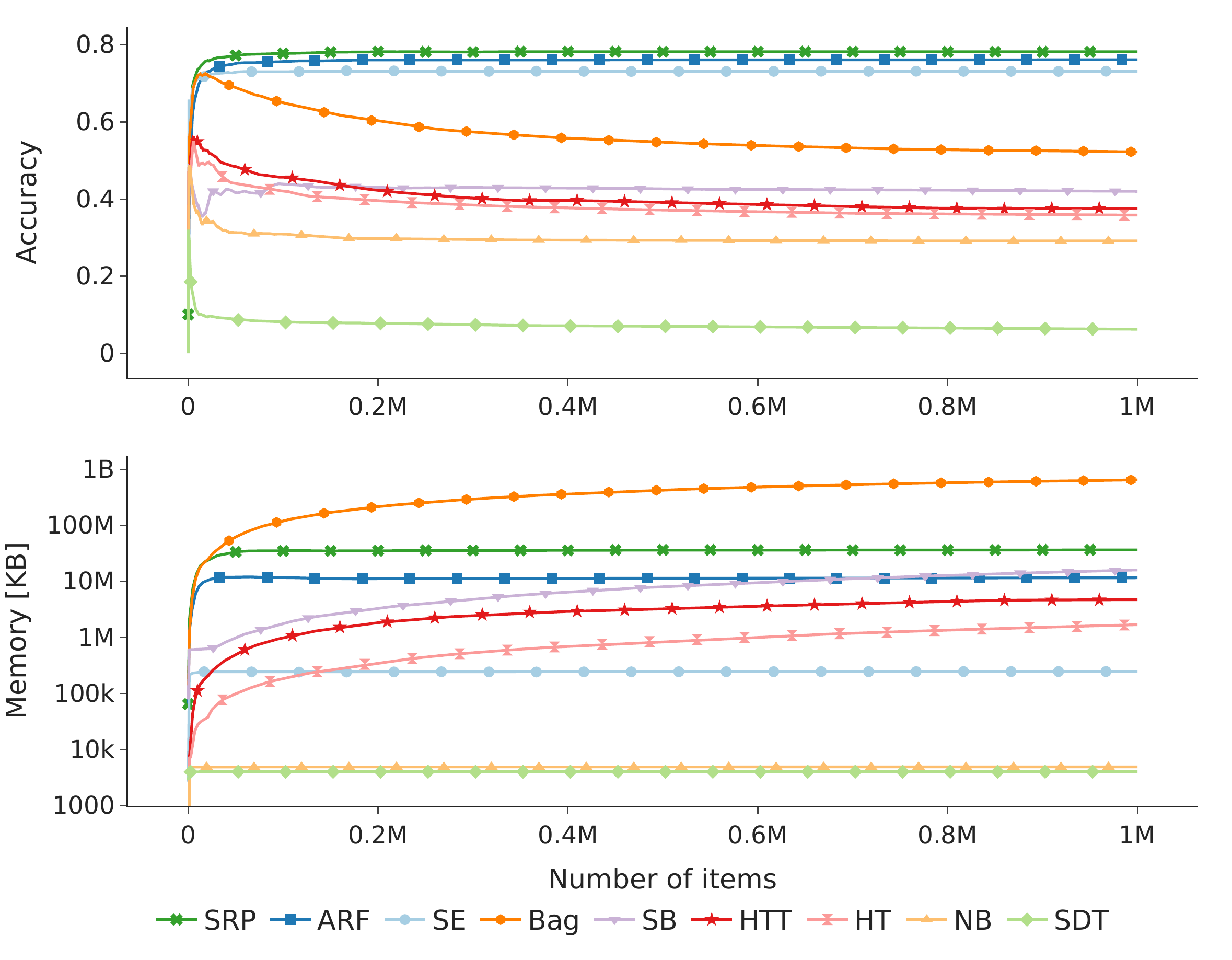}
\end{minipage}\hfill
\begin{minipage}{.49\textwidth}
    \centering 
    \includegraphics[width=\textwidth,keepaspectratio]{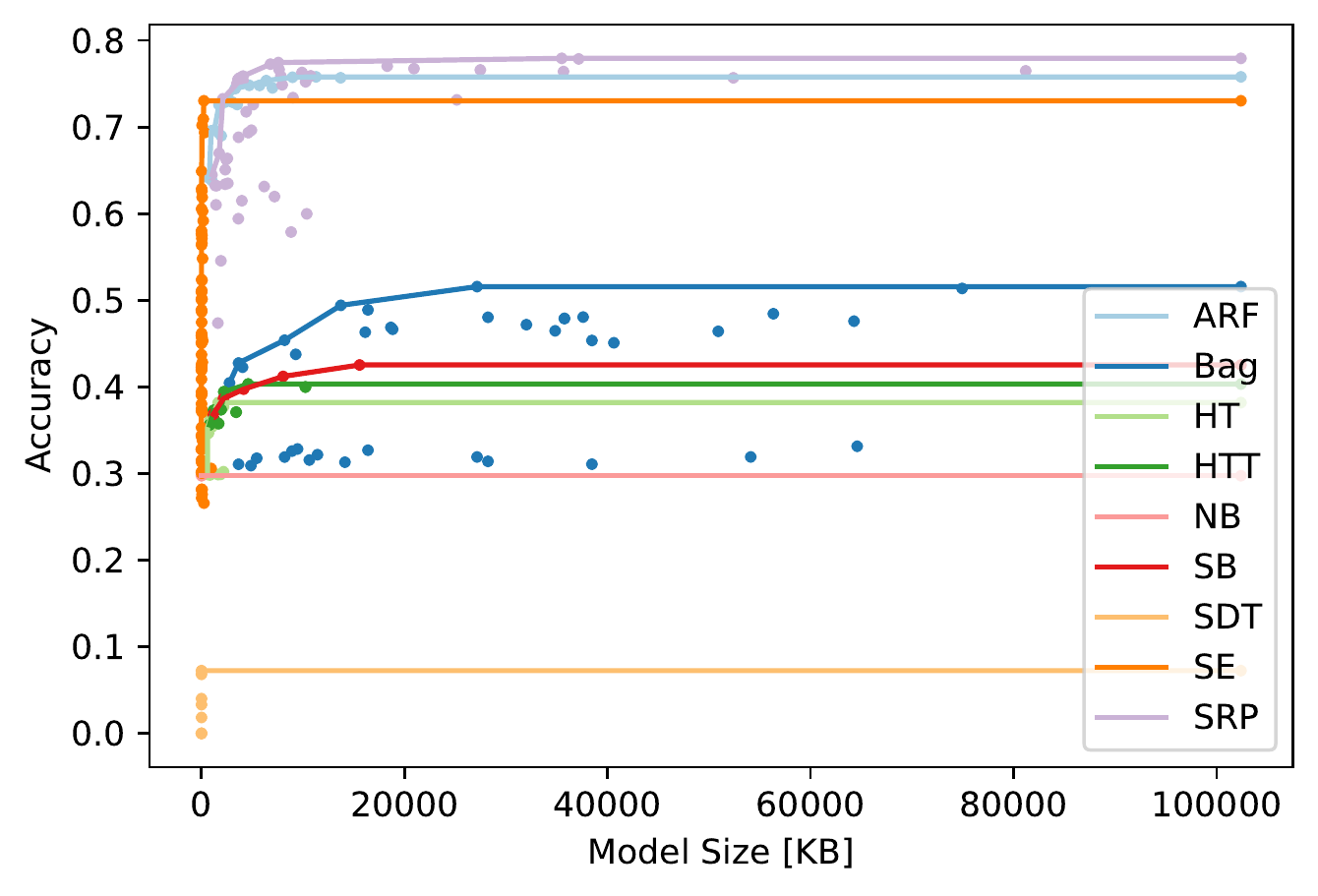}
\end{minipage}
\caption{(left) Test-then-train accuracy and memory consumption on the rbf\_f dataset of the best configuration over the number of data items in the stream. (right) Pareto front on the rbf\_f dataset of each method.}
\end{figure}

\begin{figure}[H]
\begin{minipage}{.49\textwidth}
    \centering
    \includegraphics[width=\textwidth,keepaspectratio]{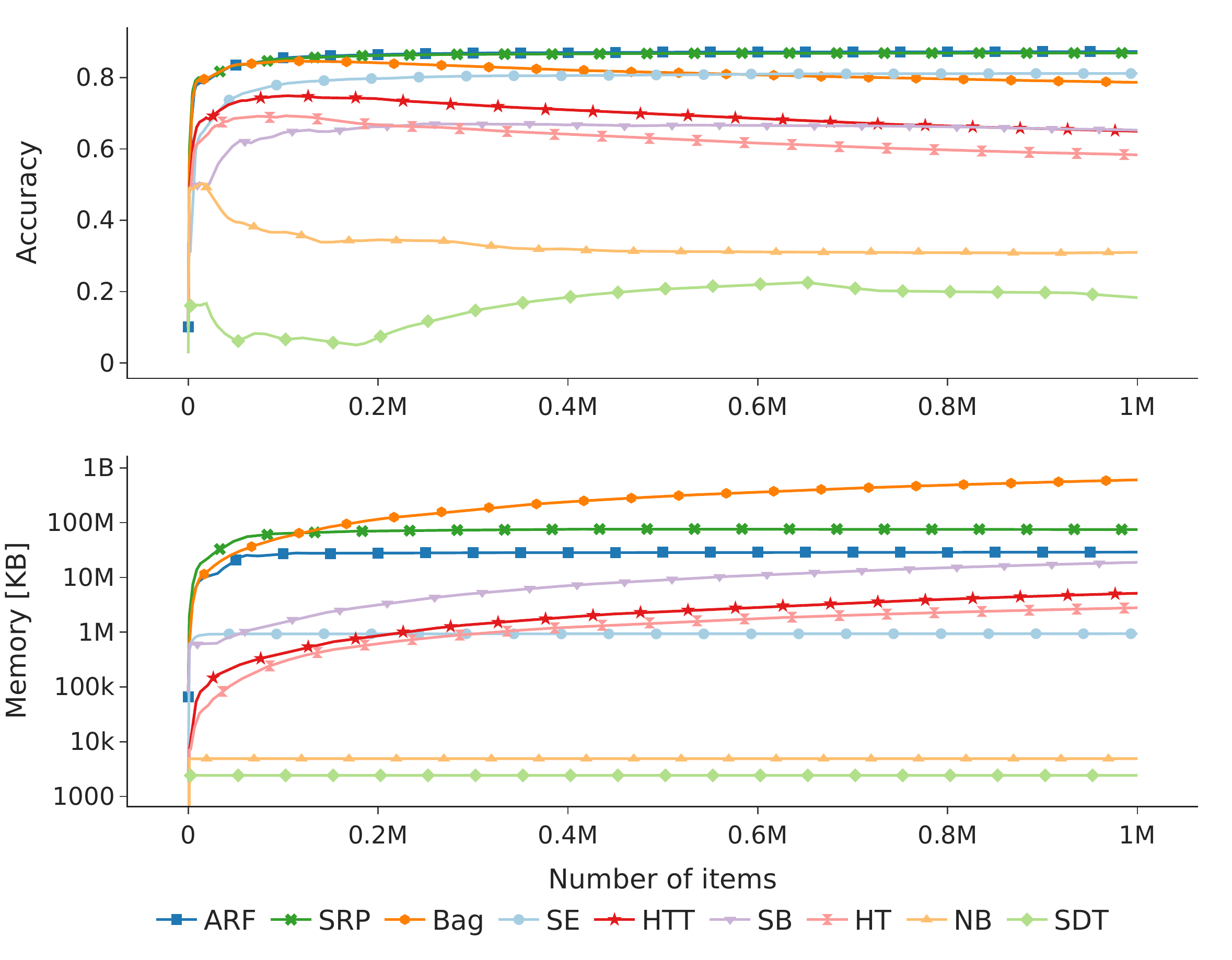}
\end{minipage}\hfill
\begin{minipage}{.49\textwidth}
    \centering 
    \includegraphics[width=\textwidth,keepaspectratio]{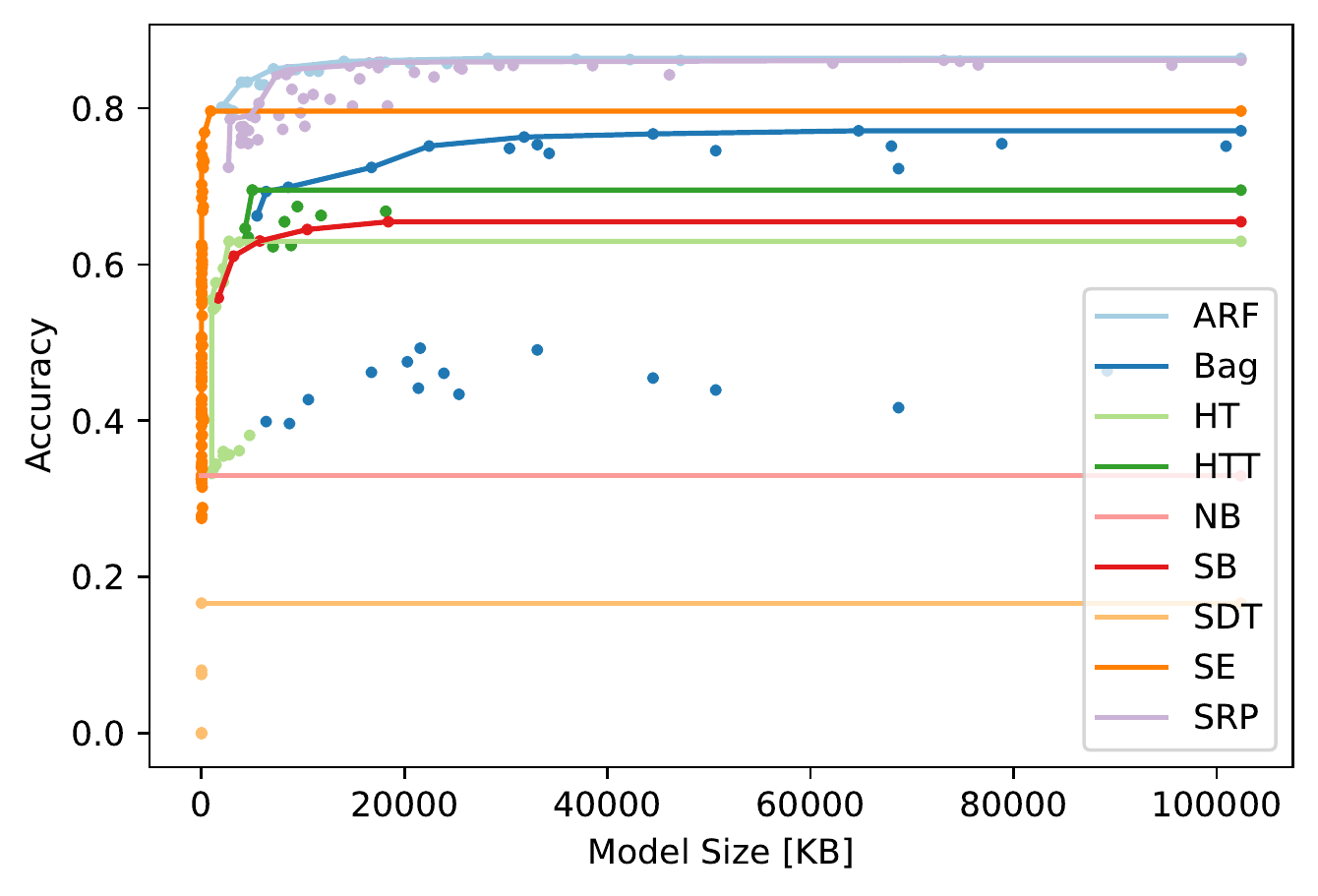}
\end{minipage}
\caption{(left) Test-then-train accuracy and memory consumption on the rbf\_m dataset of the best configuration over the number of data items in the stream. (right) Pareto front on the rbf\_m dataset of each method.}
\end{figure}

\begin{figure}[H]
\begin{minipage}{.49\textwidth}
    \centering
    \includegraphics[width=\textwidth,keepaspectratio]{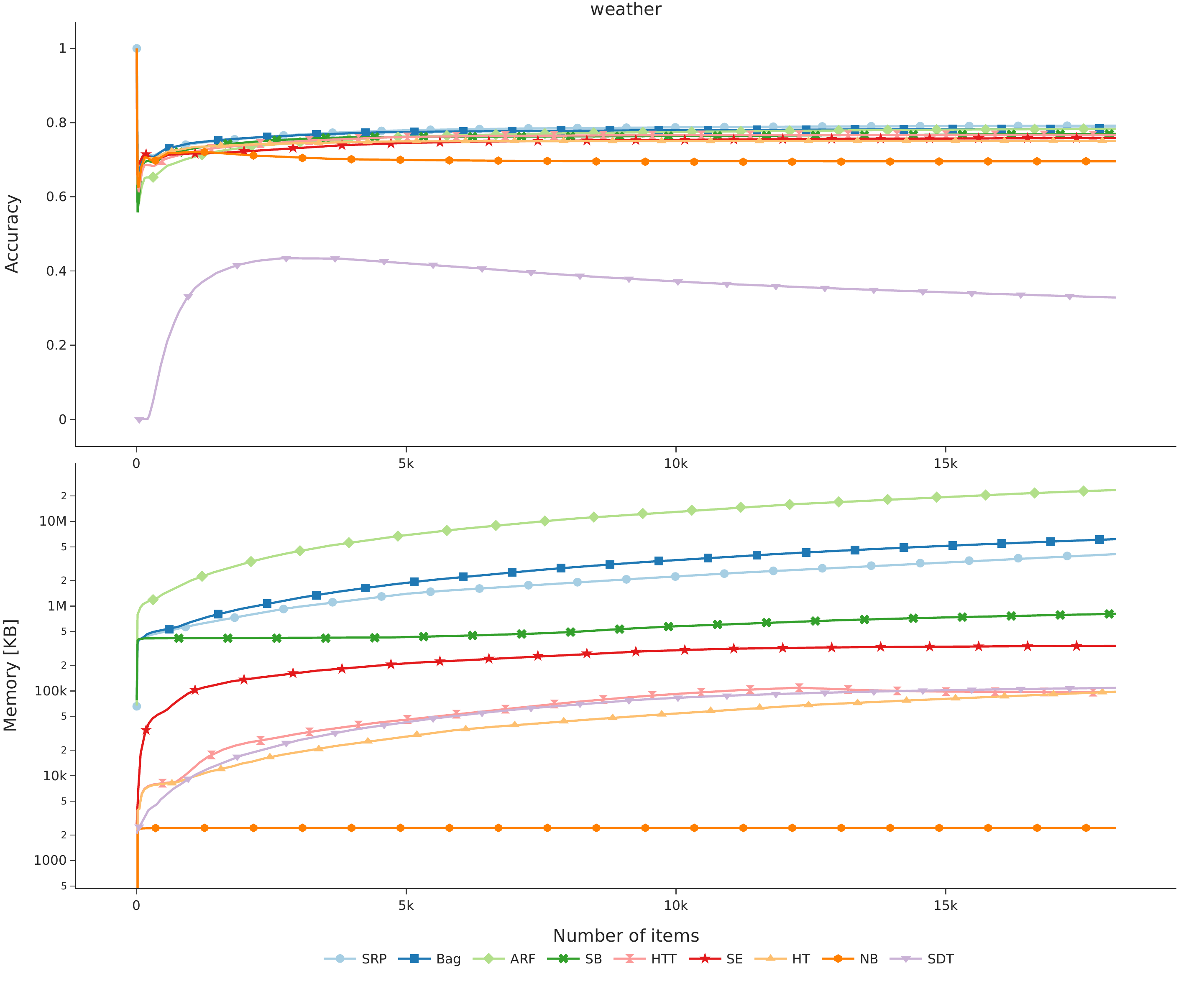}
\end{minipage}\hfill
\begin{minipage}{.49\textwidth}
    \centering 
    \includegraphics[width=\textwidth,keepaspectratio]{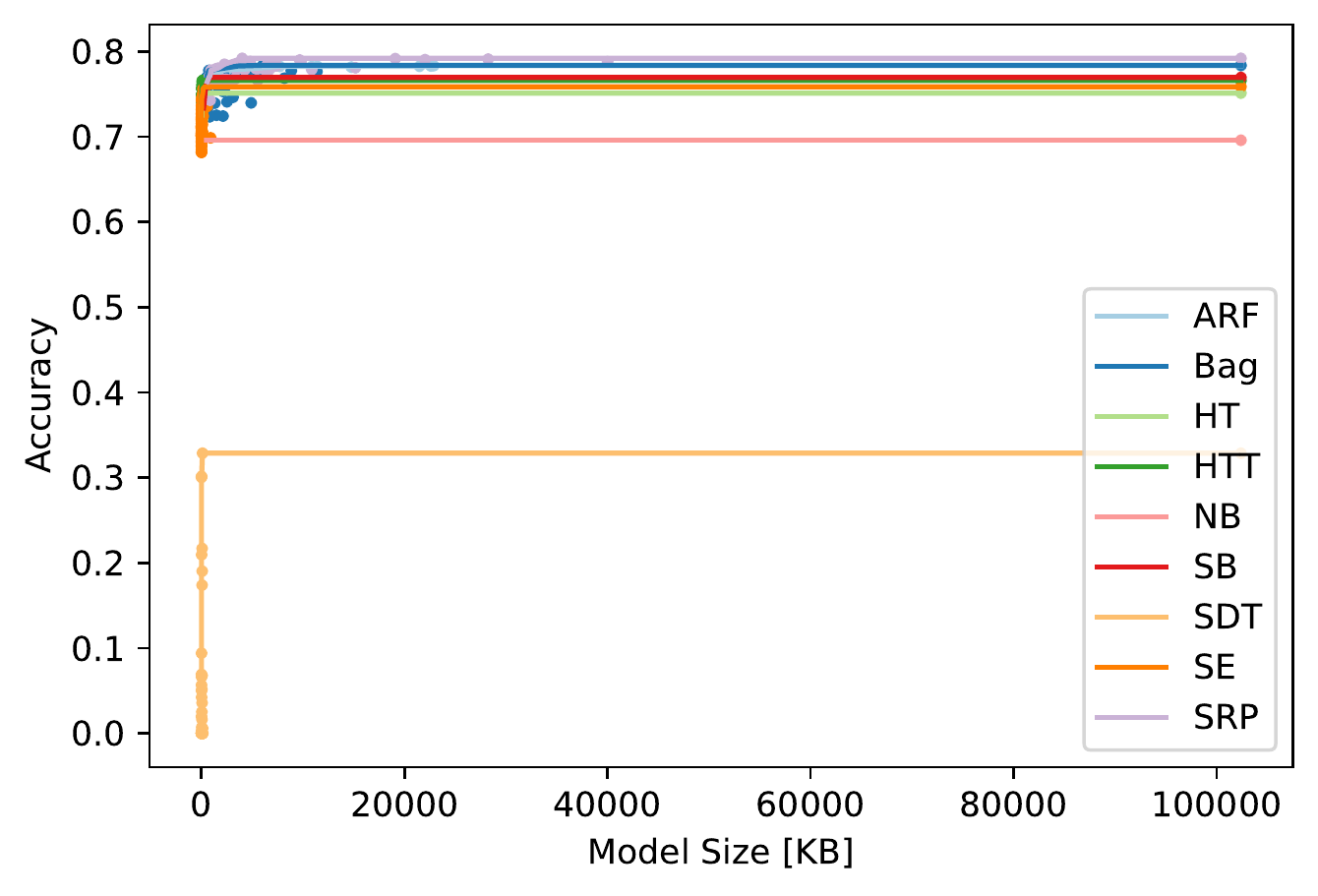}
\end{minipage}
\caption{(left) Test-then-train accuracy and memory consumption on the weather dataset of the best configuration over the number of data items in the stream. (right) Pareto front on the weather dataset of each method.}
\end{figure}

\bibliography{literatur}